\def\paperlanguage{} 

\documentclass[letterpaper, 10 pt, conference]{ieeeconf}  

\usepackage{bm}
\usepackage{cite}
\usepackage{flushend}
\include{preamble}

\usepackage{booktabs}

\IEEEoverridecommandlockouts                              

\title{\LARGE \textbf
  {
    \switchlanguage%
    {%
      Design Method of a Kangaroo Robot \\ with High Power Legs and an Articulated Soft Tail
    }%
    {%
     大出力脚と多関節構造の尻尾を有するカンガルーロボットの構成法 
    }%
  }
}

\author{Shunnosuke Yoshimura$^{1}$, Temma Suzuki$^{1}$, Masahiro Bando$^{1}$, \\Sota Yuzaki$^{1}$, Kento Kawaharazuka$^{1}$, Kei Okada$^{1}$, and Masayuki Inaba$^{1}$
  \thanks{$^{1}$ The authors are with the Department of Mechano-Informatics, Graduate School of Information Science and Technology, The University of Tokyo, 7-3-1 Hongo, Bunkyo-ku, Tokyo, 113-8656, Japan.
    {\texttt\small [yoshimura, suzuki, bando, yuzaki, kawaharazuka, okada, inaba]@jsk.t.u-tokyo.ac.jp}
  }
}
\begin{document}

\maketitle
\thispagestyle{empty}
\pagestyle{empty}

\begin{abstract}
  \switchlanguage%
  {%
    In this paper, we focus on the kangaroo, 
    which has powerful legs capable of jumping and a soft and strong tail.
    To incorporate these unique structure into a robot for utilization, 
    we propose a design method that takes into account 
    both the feasibility as a robot and the kangaroo-mimetic structure.
    Based on the kangaroo's musculoskeletal structure, 
    we determine the structure of the robot that enables it to jump 
    by analyzing the muscle arrangement and prior verification in simulation.
    Also, to realize a tail capable of body support, 
    we use an articulated, elastic structure as a tail.
    In order to achieve both softness and high power output, 
    the robot is driven by a direct-drive, high-power wire-winding mechanism, 
    and weight of legs and the tail is reduced by placing motors in the torso.
    The developed kangaroo robot can jump with its hind legs, 
    moving its tail, and supporting its body using its hind legs and tail.
  }%
  {%
    生物を規範とするロボットは、生物に固有の構造や動作に着目し、ロボットに取り入れて活用する、
    本研究では、跳躍可能で力強い脚と、柔らかさと強さを有するカンガルーに着目する。
    カンガルーを規範とした身体構造をロボットに取り入れながら、
    ロボットとしての実現可能性も考慮した等身大・大出力・柔軟なロボットの構成法を提案する。
    カンガルーの筋骨格構造を基に、筋配置の分析とシミュレーションにより跳躍を可能とするロボットの構造を決定する。
    脚の柔らかさと大出力を両立するために、ダイレクトドライブで大出力なワイヤ巻き取り機構による駆動と、モータを胴体に配置することによる軽量化を行う。
    また、身体支持が可能な尻尾の実現のために、多関節で弾性を有する劣駆動構造を利用する。
    開発したカンガルーロボットは、後脚による跳躍、尻尾の駆動、後脚と尻尾を利用した身体支持が可能である。
  }%
\end{abstract}

\section{Introduction}\label{sec:introduction}
\begin{figure}[t]
  \centering
  \includegraphics[width=1.0\columnwidth]{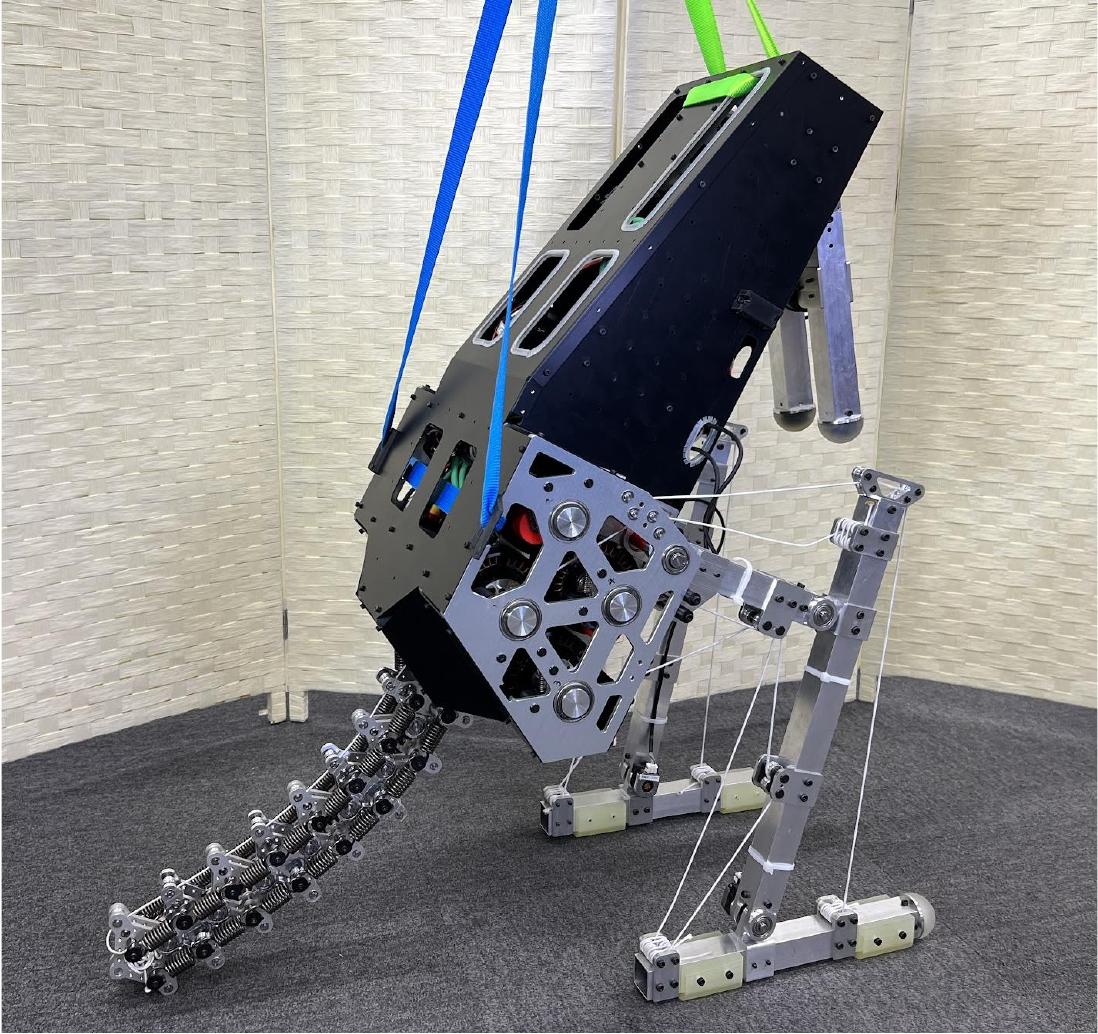}
  \vspace{-3ex}
  \caption{Overview of the kangaroo robot.}
  \label{figure:kangaroo}
  \vspace{-4ex}
\end{figure}

\switchlanguage%
{%
Bio-mimetic and bio-inspired legged robots have incorporated the unique structures and movements found in nature, 
adapting them into a form that can be utilized technically in robots.
There already exist some robots constructed based on the features of living organisms.
Rhex\cite{Rhex}, which mimics the locomotion of arthropods, 
is capable of walking on uneven terrain.
MIT Cheetah\cite{Cheetah} has a bio-mimetic spine 
that bends softly in addition to four high-powered legs.
It is capable of running utilizing its spine and legs at the same time.
RiSE\cite{RiSE} can climb walls due to its bio-inspired design.
Uniroo\cite{Uniroo} and Kenken\cite{Kenken} are
able to jump by their articulated legs based on biological structure.
Mowgli\cite{Mowgli} realized jump and landing with a pneumatic musculoskeletal system.
BionicKangaroo\cite{festo} realized leaps with energy stored in the Achilles tendon,
using the kangaroo as a norm.
There are also kangaroo robots that have made leaps 
by using its tail and legs\cite{activetail, Jun}.

To improve the function and performance of a bio-mimetic leg robot, 
it is important to realize a musculoskeletal structure 
that is life-size and unique to living organisms, with both high power and flexibility.
In this paper, we focus on the body structure of the kangaroo 
and develops a robot based on the kangaroo.
The kangaroo has strong, leapable hind legs and a tail 
that is both strong and soft enough to be used as a leg.
Although robots inspired by the kangaroo have been developed before\cite{Uniroo, festo, activetail, Jun},
creating a life-size robot based on the musculoskeletal structure of a kangaroo, 
which includes powerful legs and a tail with an articulated structure, 
still remains as a challenge.
Therefore, we propose a robot construction method that realize flexibility and high power output
by wire drive while incorporating a bio-mimetic body structure into a robot.
Based on this construction method, we build a life-size kangaroo robot, shown in \figref{figure:kangaroo},
and succeeded in letting it jump using its hind legs and tail.

In \secref{sec:method}, we describe the design method of the legs and tail
based on the kangaroo's body structure.
In \secref{sec:legs}, we describe the design of the legs.
In \secref{sec:tail}, we describe the design of the tail.
In \secref{sec:hardware}, we describe the design and implementation of the high power wire module 
and overall design of the robot.
In \secref{sec:experiments}, we perform experiments on the hind legs and tail alone 
and on their combined movements, then discuss the results of these experiments.
In \secref{sec:conclusion}, we present our conclusions 
on the design method of the kangaroo robot.
}%
{%
生物を規範とする脚ロボットは、生物が持つ固有の構造や動作に着目し、
それらを技術的に活用可能な形にしてロボットに取り入れてきた\cite{Rhex, Cheetah, RiSE, Uniroo, Kenken, Mowgli, festo, activetail, Jun}。

これまで開発されてきた生物規範型脚ロボットを挙げる。
節足動物のロコモーションを模倣したRhex\cite{Rhex}では、リンク機構を活用した6本の脚を利用して
不整地歩行を可能にしている。
MIT Cheetah\cite{Cheetah}は、大出力の四脚に加えて柔らかに曲がる生物模倣型の背骨を有する。
背骨と脚を同時に活用した走行を可能にした。
RiSE\cite{RiSE}は、生物をもとにした設計により壁を登ることができる。
Uniroo\cite{Uniroo}やKenken\cite{Kenken}は、生物をもとにした多関節型の脚により跳躍を実現した。
Mowgli\cite{Mowgli}は、空気圧による筋骨格系で跳躍と着地を実現した。
BionicKangaroo\cite{festo}は、カンガルーを規範として、アキレス腱にエネルギーを蓄えた跳躍や尻尾を実現した。
尻尾や脚の利用により跳躍を行ったカンガルーロボット\cite{activetail, Jun}も存在する。

生物規範型脚ロボットの機能・性能の向上には、
生物が有する、大出力と柔軟さを両立した、等身大で生体に固有の筋骨格構造を実現することが
重要である。
そこで、本研究ではカンガルーの身体構造に着目し、カンガルーを規範としたロボットを開発する。
カンガルーは、強力で跳躍可能な後脚と、脚としても利用できる強さと柔らかさを有する尻尾を持つ。
これまでにもカンガルーを規範としたロボットは存在した\cite{Uniroo, festo, activetail, Jun}が、
このような等身大筋骨格構造の実現は課題となっている。
そこで、生体規範型の身体構造をロボットに取り入れながら、
ワイヤ駆動による柔軟さ・大出力といった性能を実現するロボット構成法を提案する。
その構成法をもとに、等身大のカンガルーロボットを製作し、
後脚と尻尾を利用した動作を行わせる。
開発したカンガルーロボットを\figref{figure:kangaroo}に示す。

\secref{sec:method}ではカンガルーの身体構造を規範とする脚および尻尾の構成法を述べる。
\secref{sec:legs}では脚の設計について述べる。
\secref{sec:tail}では尻尾の設計について述べる。
\secref{sec:hardware}では大出力ワイヤモジュールおよび、ロボット全体の設計と実装を述べる。
\secref{sec:experiments}では尻尾・後脚の単体実験とそれらを組み合わせた動作実験を行い、その実験結果について議論する。
\secref{sec:conclusion}では実験結果を踏まえてカンガルーロボットの構成法について結論を述べる.
}%

\section{design method of a kangaroo robot} \label{sec:method}
\begin{figure}[t]
  \centering
  \includegraphics[width=1.0\columnwidth]{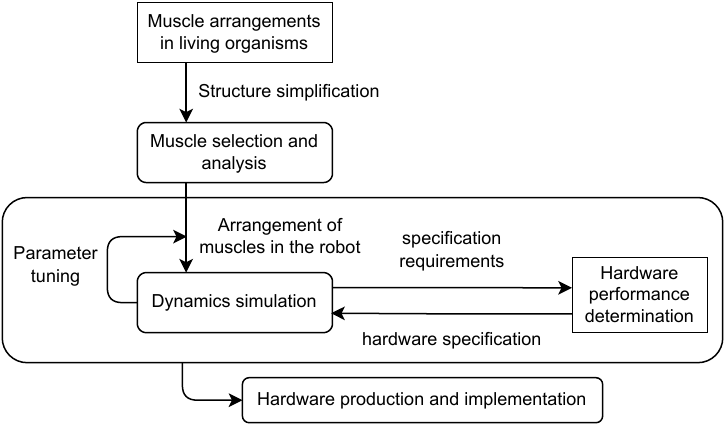}
  \vspace{-3ex}
  \caption{Design procedure for life-size kangaroo robot.}
  \label{figure:design_process}
  \vspace{-3ex}
\end{figure}

\switchlanguage%
{%
In this paper, we realize a life-size robot with legs and a tail 
based on the body structure of a kangaroo and make them to perform high-power and flexible movements.
The design concept of the life-size musculoskeletal robot based on the kangaroo is as follows. 
First, the biological structure of a kangaroo, such as its muscle arrangement and skeletal structure, 
is used as the norm. 
Second, trade-offs among performance, quantity and weight of motors, etc are required to be balanced in designing the life-size robot, 
in order to achieve kangaroo's high-power moion as well as its flexibility.

Thus, in designing a life-size musculoskeletal kangaroo robot, 
it is necessary to consider both the conditions based on the kangaroo body structure and its feasibility as a robot.
\figref{figure:design_process} shows the design process that takes them into account. 
First, the structure used in the living body is used as the base musculoskeletal structure. 
Next, the musculoskeletal structures are selected and analyzed.
The number of muscles need to be limited as an implementation requirement for a wire-driven robot. 
In particular, for areas such as the legs where high-output movements are required, 
the use of selected muscles is analyzed based on the trajectory of the movement.
Then, additional requirements for wire drive implementation are considered and a feasible muscle arrangement is created.
The motion considering actual control is simulated to confirm that 
the motion is feasible with the body structure and actuator performance.
In parallel, hardware selection and performance decisions are made. 
In particular, the choice of motors and reduction ratios are important, 
because they determine the performance of the leg robot hardware. 
The higher the output and torque of the leg robot's hardware is, 
usually the greater the weight of the robot's body is, 
and the greater the performance required is.
Therefore, the performance requirements based on the living body analysis and 
the performance of the hardware must be well matched. 
The kangaroo robot is designed according to the above procedure. 
}%
{%
本研究では、
カンガルーの身体構造を規範とした、脚と尻尾を持つ等身大のロボットを実現し、
大出力・柔軟動作を行う。
カンガルーを規範とする等身大筋骨格ロボットの設計思想は次のようになる。
まず、カンガルーの筋配置や骨格構造といった生体構造を規範とする。
ただし、等身大ロボットを製作する上では、
モータの性能や重量・個数の制約およびトレードオフ、
ハードウェアとしての設計・実装上の条件が存在する。
それらを踏まえた上で、カンガルーの大出力動作と柔軟さを実現する必要がある。

このように、カンガルーを規範とする等身大筋骨格ロボットの設計においては、
生体規範的な条件と、ロボットとしての実現可能性の両方を考慮した設計を
行う必要がある。
そこで、これらを考慮した設計プロセスを\figref{figure:design_process}に示す。
まず、生体の筋骨格構造を考察する。
解剖学を考慮し、生体において利用されている構造をベースの筋骨格構造とする。
次に、筋骨格構造の選択と解析を行う。
ワイヤ駆動ロボットといて製作する上での実装要件として、筋を限定する必要がある。
特に、脚のように大出力動作が必要な場合は、
選択した筋がどのように利用されているのかを、
動作の軌道から解析する。
そして、動作シミュレーションを行う。ワイヤ駆動の実装要件をさらに追加し、
実装可能な筋配置を作成する。
実際の制御を考慮した動作をシミュレーションし、
定めた構造およびアクチュエータの性能で動作が可能であることを確かめる。
これらと並行して、ハードウェアの選択や性能決定を行う。
特に、モータの選択や減速比の決定が重要となる。
なぜなら、脚ロボットのハードウェア性能において、
出力やトルクが大きいほど、身体の重量も増えて、動作に必要な性能が大きくなるといった
トレードオフの関係が存在する。
よって、ロボットの要求性能とハードウェアの性能の相性が良い必要があるからである。

以上のような手順によりカンガルーロボットを設計する。
以下、脚と尻尾のそれぞれについて各手順を行う。
}%

\section{Design of legs} \label{sec:legs}
\subsection{Musculoskeletal Structure of Living Kangaroo Leg}
\begin{figure}[t]
  \centering
  \includegraphics[width=1\columnwidth]{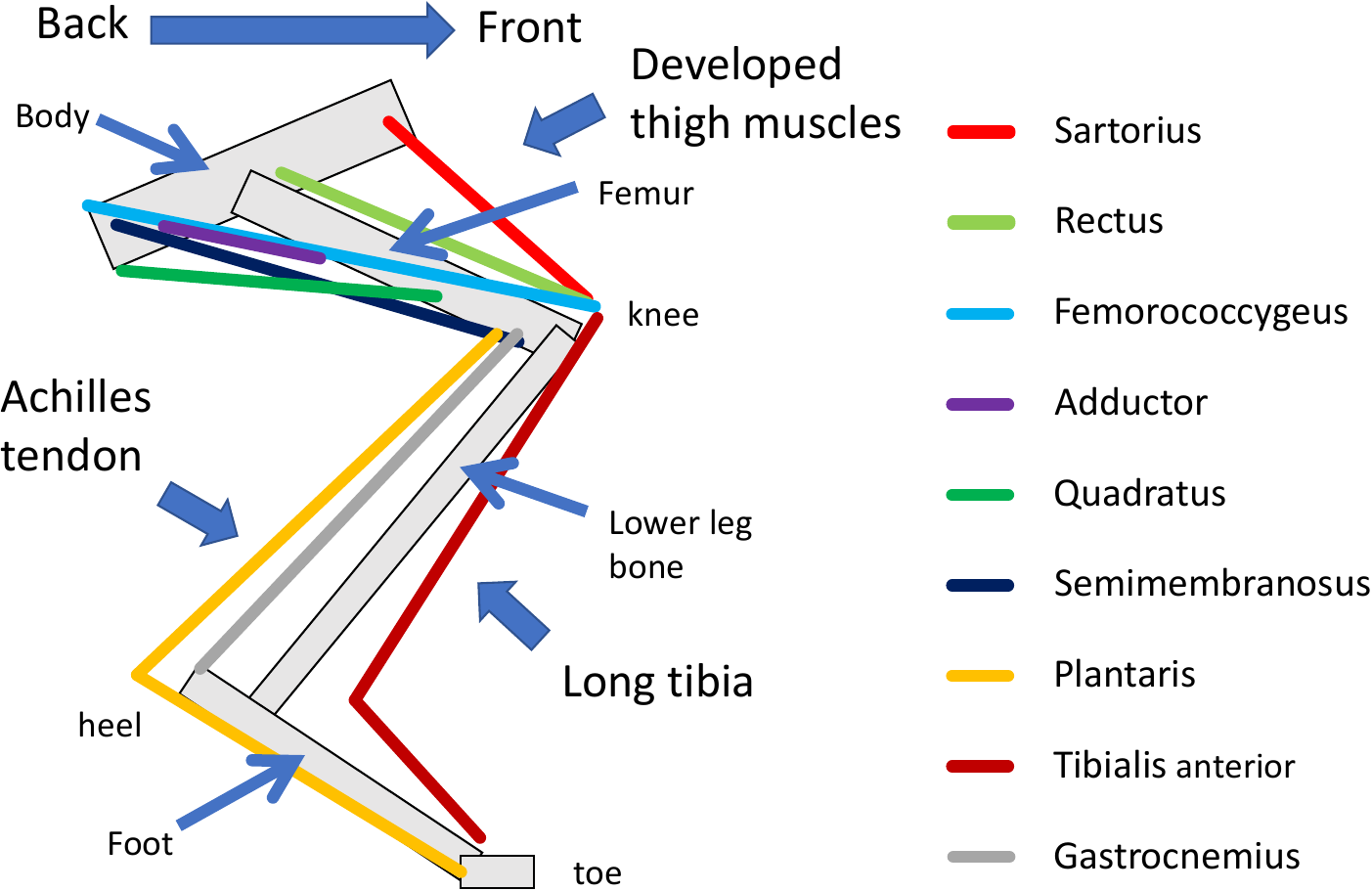}
  \vspace{-3ex}
  \caption{Hind leg structure and typical muscles contributing to kangaroo leg movement.}
  \vspace{-3ex}
  \label{figure:anatomy_leg}
\end{figure}

\switchlanguage%
{%
We describe the musculoskeletal structure of the kangaroo based on anatomy\cite{Mechanicsofkangaroo}. 
First, the muscles in the hind legs that contribute to leaping are shown in \figref{figure:anatomy_leg}. 
In the kangaroo's hind leg, the bones of the lower leg are longer than those of the thigh and foot. 
Also, at the heel, the bone protrudes from the joint. 
This is thought to increase the moment arm and the contribution of muscles to joint torque. 
The thigh muscles and sartorius muscles that connect the upper knee to the torso are well-developed. 
The thigh muscles are considered to play a major role in walking and jumping. 
In addition, the gastrocnemius muscle, one of the muscles that connect the Achilles tendon, is located from the heel to the thigh. 
This elastic energy is believed to be used during jumping\cite{Mechanicsofkangaroo}.
}%
{%
解剖学\cite{Mechanicsofkangaroo}に基づくカンガルーの筋骨格構造について述べる。
まず、後脚について、跳躍をはじめとする
脚を利用した運動に寄与する筋肉を\figref{figure:anatomy_leg}に示す。
カンガルーの後脚は、下腿部の骨が大腿部および足部と比べて長くなっている。
また、踵では、骨が関節部分から出っ張っている。
これにより、モーメントアームを大きくして筋肉の関節トルクへの寄与を大きくしていると
考えられる。
特に発達しているのは膝上部から胴体に繋がる大腿筋や縫工筋などの筋肉である。
大腿筋は歩行や跳躍において大きな役割を果たすとされる。
また、アキレス腱を成す筋肉のひとつである腓腹筋が踵から大腿部に向けて配置しており、
跳躍時はこの弾性エネルギーを利用しているとされる。
}%

\subsection{Muscle Selection and Analysis}
\begin{figure}[t]
  \centering
  \includegraphics[width=0.5\columnwidth]{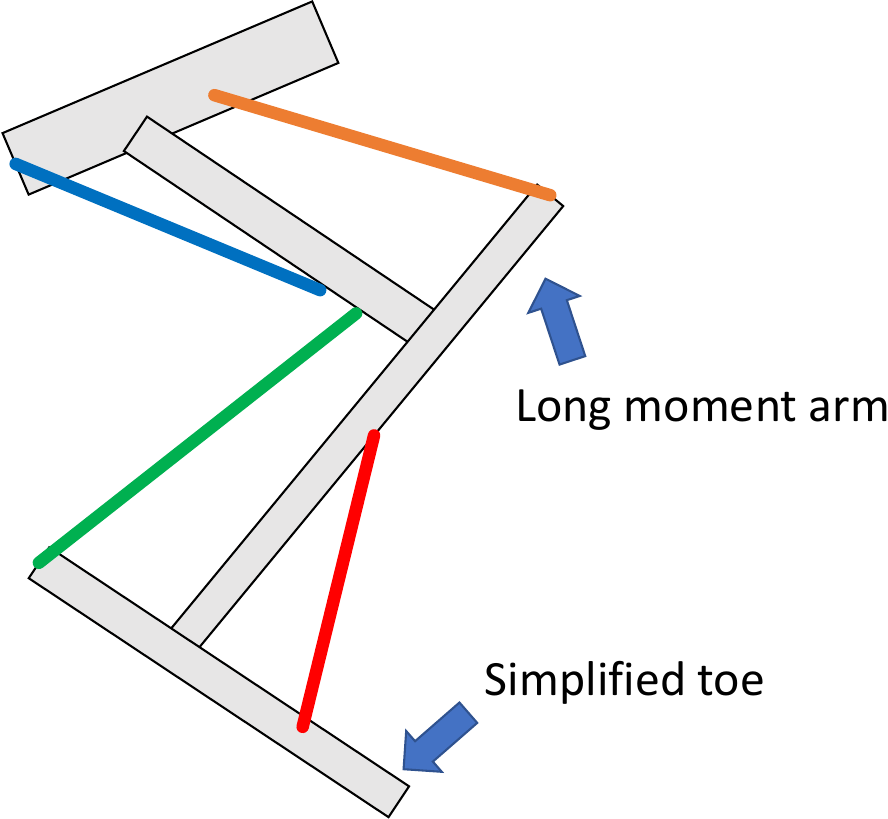}
  \vspace{-1ex}
  \caption{Simplified musculoskeletal structure of the leg.}
  \vspace{-2ex}
  \label{figure:simplified_leg}
\end{figure}

\begin{figure}[t]
  \centering
  \includegraphics[width=0.6\columnwidth]{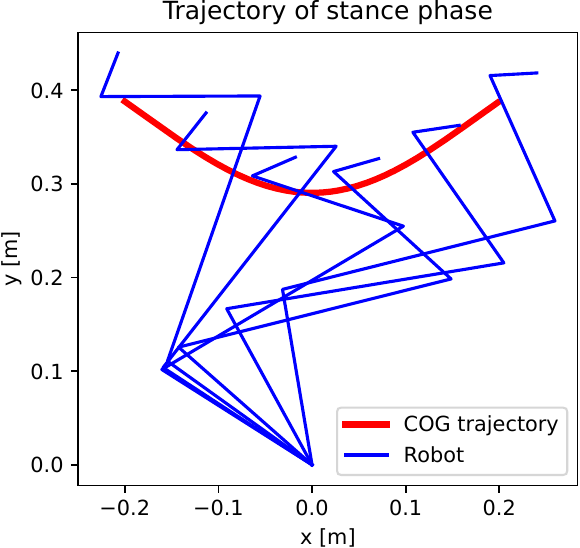}
  \vspace{-2ex}
  \caption{Trajectory of the stance phase. 
  The blue broken line represents the robot's body to its legs, with the bent parts indicating the joints.
  From top to bottom, it represents the torso, thigh, lower leg, and footplate of the robot.
  }
  \vspace{-3ex}
  \label{figure:jump_trajectory}
\end{figure}

\begin{figure}[t]
  \centering
  \includegraphics[width=1\columnwidth]{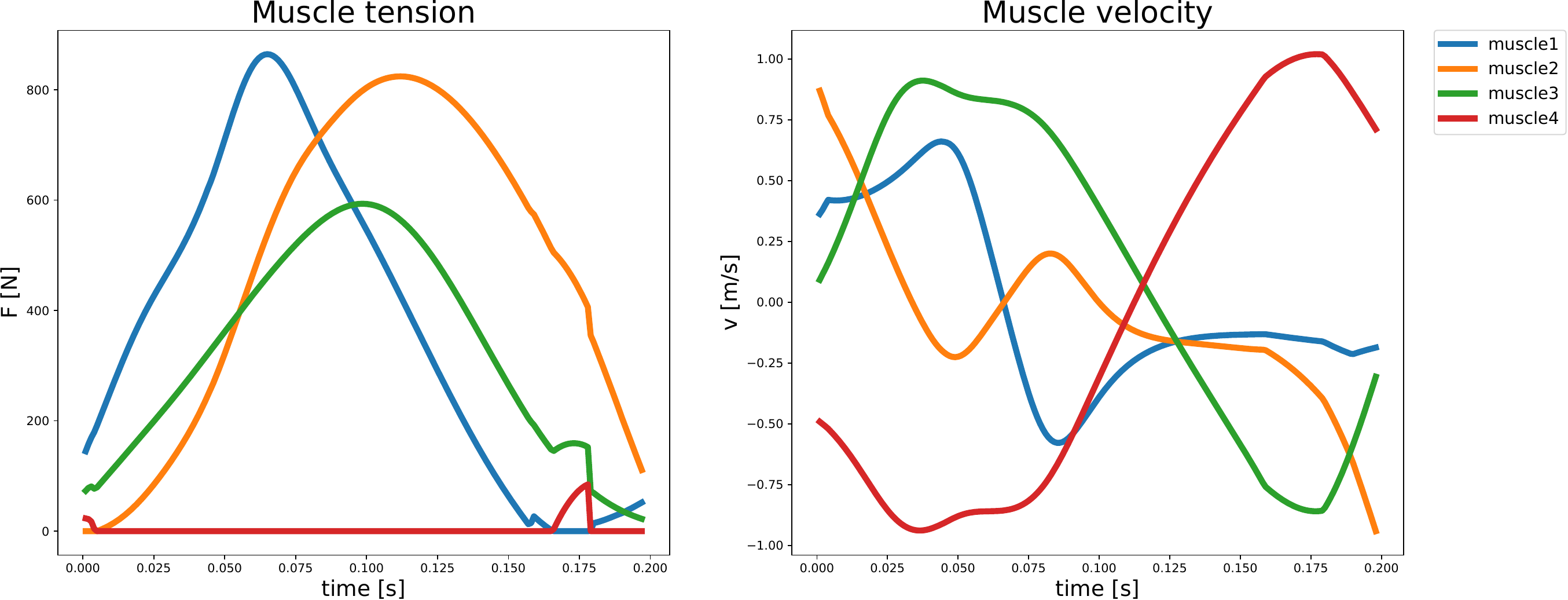}
  \vspace{-2ex}
  \caption{Muscle tension and velocity during stance phase. 
  The color of the plot is the same as the color of the muscles in \figref{figure:simplified_leg}.}
  \vspace{-2ex}
  \label{figure:muscle_trajectory}
\end{figure}

\switchlanguage%
{%
The musculoskeletal structure based on the living organisms has many muscles and is difficult to implement directly into a robot.
Hence, we analyzed the use of muscles in movements after simplifying the musculoskeletal structure. 
In simplifying the structure, it is necessary to select a lightweight structure 
with a small number of actuators while enabling jumping and body support. 
Therefore, a simplified musculoskeletal structure is shown in \figref{figure:simplified_leg}, 
which is a two-dimensional leg with three rotational joints and four muscles. 
The toes are assumed to be capable of both point and foot surface placement. 
Two hind legs are provided on the robot. 
In the design phase, one leg is modelized and its dynamics are calculated to obtain the required muscle forces and placement. 
The required muscle performance is calculated by considering the trajectories of the legs and torso in the case of jumping, 
where the rear legs would require the most force. 
The following procedure is used.

\begin{enumerate}
  \item Calculate the trajectory of a leap.
  \item Calculate joint torque by inverse dynamics.
  \item Calculate muscle tension and velocity.
\end{enumerate}

First, the trajectory of the leap is determined. 
Especially in continuous jumping, the kangaroo's body forms a stable trajectory in which the stance phase and the flight phase appear alternately. 
For simplicity, we obtain a center of gravity (COG) trajectory in which the horizontal COG velocity is constant, 
the vertical COG trajectory is sine wave in the stance phase, and parabolic in the flight phase. 
This trajectory is obtained by determining the horizontal velocity and distance of COG during the stance and flight phases, 
and by considering the boundary condition that the position and velocity are the same when the phases are interchanged. 
This COG trajectory is divided, and inverse kinematics is solved to obtain the joint angle trajectory in jumping. 
The physical parameters used are shown in \tabref{table:traj_table}. 
The joint angle trajectory during the stance phase, which especially requires a large force, is shown in \figref{figure:jump_trajectory}.

Next, numerical differentiation is performed for the joint angle trajectory, 
and the joint torque is calculated from the inverse dynamics. 
The hind leg is assumed to be point-grounded at the toes. 
If the torque is calculated as a serial link, the toe torque does not necessarily become zero and the motion satisfying the trajectory cannot be realized. 
Therefore, the torque is calculated assuming that the trajectory can be realized by considering the effect that the torso receives from the tail. 
In other words, the toe torque is set to 0, and the torque and force received by the body from the tail are added instead. 
Letting $\bm{f}_{tail}$ be the torque and force received from the tail and $J_{tail}$ be the Jacobian of the tail's effect on the body. 
The equation of motion is as follows.
\begin{eqnarray}
  M(\bm{\theta}) \ddot{\bm{\theta}} + \bm{b}(\bm{\theta}, \dot{\bm{\theta}}) = 
  \begin{bmatrix}
    0 \\ \bm{\tau}'
  \end{bmatrix}
   + J_{tail}^T \bm{f}_{tail}
\end{eqnarray}
The joint torques except for the toes are represented by $\bm{\tau}'$.
When the left side is calculated from the joint angle trajectory, 
this equation becomes a simultaneous linear equation for the torque on the right side, and the torque can be calculated.
Note that if the number of divisions of the joint angle orbit is not sufficiently large, the error in numerical differentiation will be large.

Finally, we calculate muscle tension and muscle velocity.
If $\theta,L$ is the joint angle and muscle length, the muscle length Jacobian $G$ is defined as follows.
\newcommand{\pa}[2]{\frac{\partial {#1}}{\partial {#2}}}
\begin{eqnarray}
  G(\theta) &=& \pa{L}{\theta}
\end{eqnarray}
Let $\bm{\tau}, \bm{f}$ be the joint torque and muscle tension.
The conversion from joint torque to muscle tension is performed by solving an optimization problem in the following equation.
\begin{eqnarray}
  \text{minimize} && || \bm{f} ||^2 \\
  \text{subject to} &&
  \left\{
    \begin{array}{ll}
      \bm{\tau} = -G^T  \bm{f} \\
      \bm{0} \leq \bm{f} \leq \bm{f_{max}}
    \end{array}
  \right\}
\end{eqnarray}

The muscle tension obtained by the above method and the muscle velocity obtained 
by numerical differentiation of the muscle length are shown in \figref{figure:muscle_trajectory}.

In the hypothetical musculoskeletal structure determined above, 
three of the four muscles have maximum muscle tension of approximately 600 to 800 N, 
and these three muscles contribute to movement almost equally. 
In addition, all of them have a maximum muscle velocity of about 1 m/s. 
On the other hand, the remaining muscle has almost zero muscle tension and does not contribute to movement.
}
{%
}

\begin{table}[tbhp]
 \begin{center}
  \caption{Physical parameters used in analysis by motion trajectory}
  \begin{tabular}{cc} \toprule
  Parameter & Value \\ \midrule
  Body mass & 14 kg \\
  Leg link1 mass & 0.3 kg \\
  Leg link2 mass & 0.5 kg \\ 
  Leg link3 mass & 0.4 kg \\ 
  Horizontal COG velocity & 2 m/s \\
  Distance of horizontal COG shift in the stance phase & 0.4 m \\
  Distance of horizontal COG shift in the flight phase & 0.6 m \\
  \bottomrule
  \end{tabular}
  \label{table:traj_table}
 \end{center}
\vspace{-1.5ex}
\end{table}

\subsection{Jump Dynamics Simulation}
\begin{figure}[t]
  \centering
  \includegraphics[width=0.6\columnwidth]{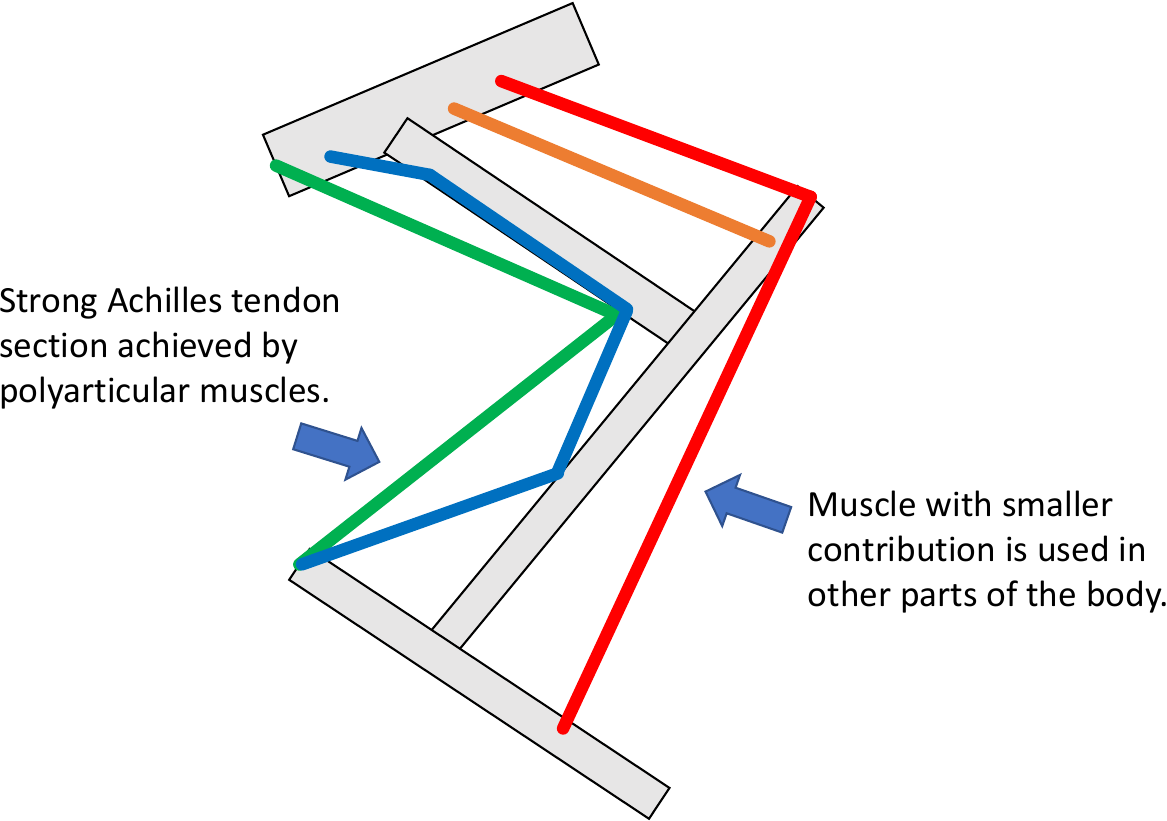}
  \vspace{-1ex}
  \caption{Leg structure using polyarticular muscles.}
  \vspace{-3.5ex}
  \label{figure:leg_use}
\end{figure}

\begin{figure}[t]
  \centering
  \includegraphics[width=1.0\columnwidth]{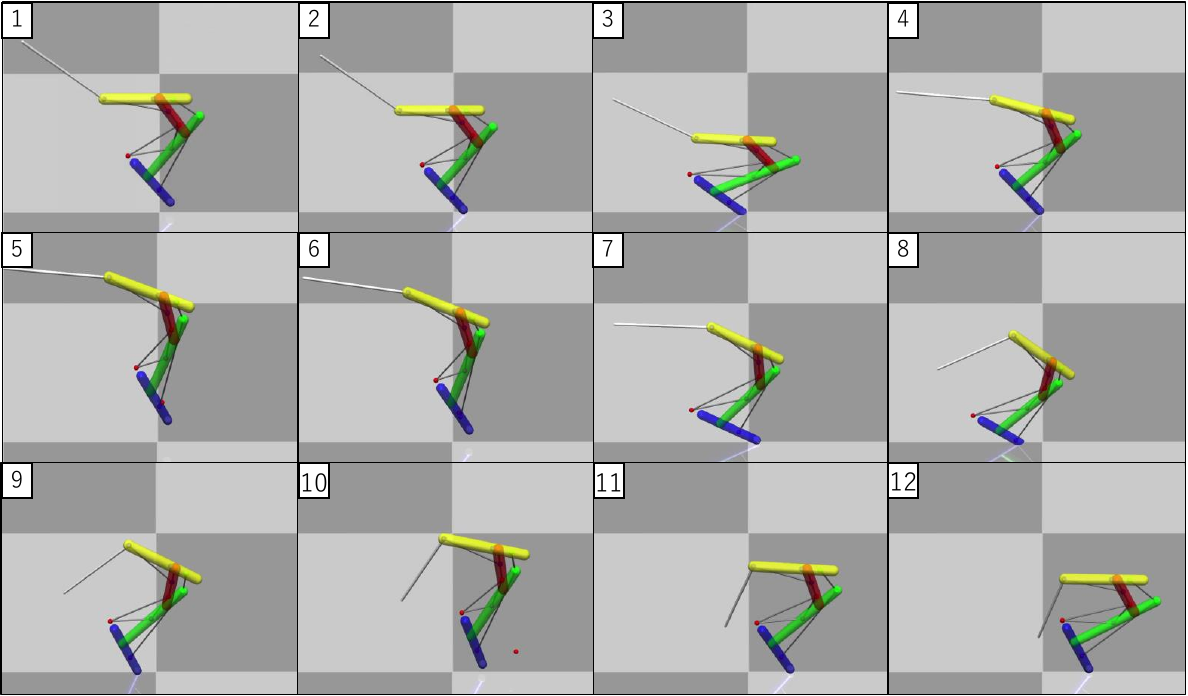}
  \vspace{-2.8ex}
  \caption{Jump simulation in mujoco.}
  \vspace{-1ex}
  \label{figure:mujoco_jump}
\end{figure}

\begin{figure}[t]
  \centering
  \includegraphics[width=1\columnwidth]{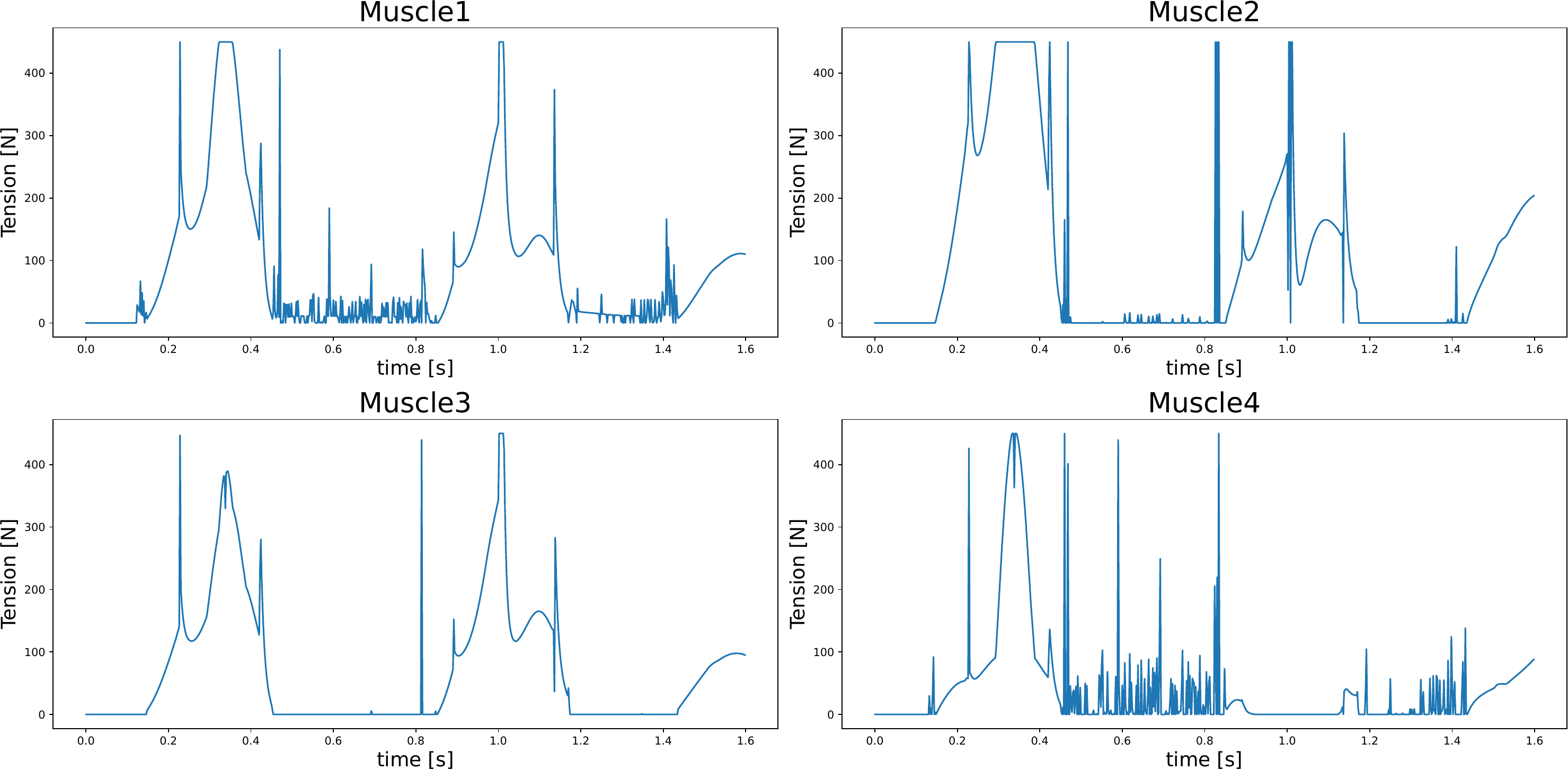}
  \vspace{-4ex}
  \caption{Muscle tension in jump simulation.}
  \vspace{-4ex}
  \label{figure:mujoco_muscles}
\end{figure}

\begin{figure}[t]
  \centering
  \includegraphics[width=1\columnwidth]{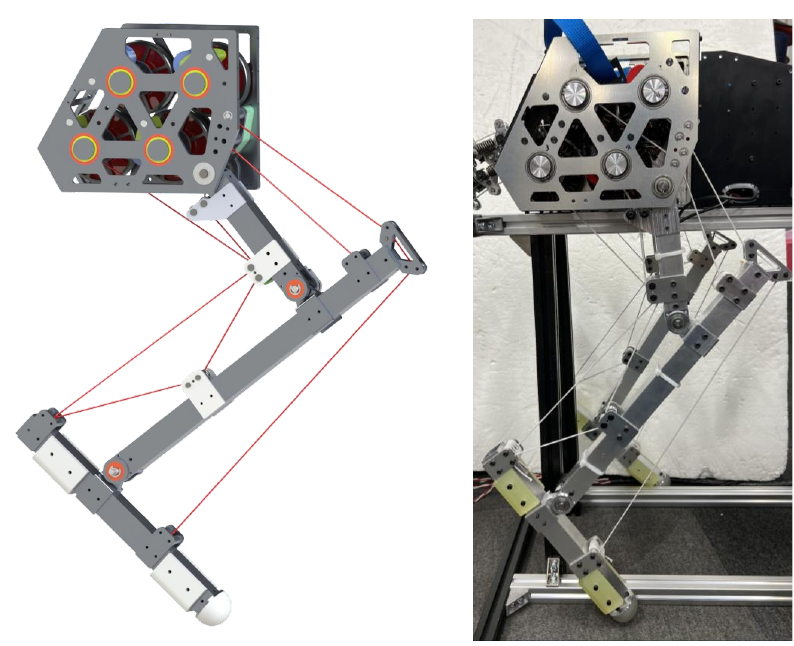}
  \vspace{-4ex}
  \caption{Mechanical design of the legs.}
  \vspace{-1ex}
  \label{figure:leg_design}
\end{figure}

\switchlanguage%
{%
In wire-driven robots, the wire can be extended to the tip and driven by wire relay points or pulleys without placing the motor in the moving part. 
Therefore, all actuators are placed in the body part, 
considering the improvement of movement performance by reducing the weight of the legs and the difficulty of placing actuators in the legs. 
However, this increases the number of wires and actuators. 
This also increases the weight, which adversely affects movement performance 
unless the contribution of the muscles to the movement is increased. 
In addition, a more complicated wire arrangement is required to achieve a torque space that allows various movements with a small number of wires.
Therefore, the goal is to design a structure in which fewer actuators can all contribute to the motion, 
while prioritizing the ability to provide the necessary torque in jumping and body support. 
Thus, based on the currently selected muscle arrangement, it is necessary to create an arrangement 
in which all muscles extend from the torso and can contribute to jumping with a small maximum muscle tension. 
In addition, it is not necessary for the COG trajectory to be sine wave in the actual jumping. 
The ideal muscle arrangement and control should be such that jumping can be performed with small muscle tension and work. 
Therefore, we will examine the muscle arrangement by dynamics simulation.

Dynamics simulations were conducted to see if the selected muscle arrangement would allow the robot to jump under control. 
Mujoco was used as the physics engine.  
In addition to the four links consisting of the torso and legs, a simplified two-dimensional model with the tail is attached to the torso as one link. 
The control of the hopping motion is based on the control algorithm by Raibert\cite{Raibert}. 
First, in the stance phase, the force between the robot's COG and the toe, which is a virtual spring force, is determined. 
The joint torques and muscle tensions that can exert these forces are calculated. 
In addition, since the amplitude of hopping decreases due to the loss of energy when virtual spring force alone is used, 
Energy-Shaping Control is used to maintain the height, similar to the controller of RAMIEL\cite{RAMIEL}. 
Also, during the flight phase, the legs are moved to the toe position target. 
This position target is determined by the CG-print\cite{Raibert}, which maintains a constant horizontal velocity during the stance phase, 
and by feedback on the target velocity.

Based on the simplified muscle arrangement, a muscle arrangement that extends to the torso and utilizes articulated muscles is shown in \figref{figure:leg_use}. 
The simulation is shown in \figref{figure:mujoco_jump}. 
This muscle arrangement was obtained experimentally and allows several jumps. 
Muscle tensions are shown in \figref{figure:mujoco_muscles}. 
In this arrangement, the areas requiring greater force are supplemented by the articulated muscles. 
This makes it possible to utilize all four muscles to jump at a maximum tension of 450 N.

The mechanical design of the legs using the muscle arrangement established above is shown in \figref{figure:leg_design}. 
It has four wire modules in the torso and wires are routed to the leg by wire relay points.
}%
{%
ワイヤ駆動においては、モータを可動部分に配置せずに、
経由点やプーリーによってワイヤを先端へと伸ばして駆動できる。
そこで、脚部の軽量化による運動性能の向上や、
アクチュエータの脚部配置の困難さを考慮して、
アクチュエータを全て胴体部分に配置する。
但し、その場合ワイヤとアクチュエータの数が増える。
これによって重量が大きくなり、
跳躍等の運動への寄与を大きくしない限り運動性能に悪影響を及ぼす。
また、少ないワイヤ数で様々な動作が可能なトルク空間を実現するためには、
より煩雑なワイヤ配置が必要となる。
そこで、跳躍や身体支持において必要なトルクを発揮できることを優先しつつ、
少ないアクチュエータが全て運動に対して寄与できる
構造の設計を目指す。
従って、現在選択した筋配置を基として、全ての筋が胴体から伸び、
かつ小さい最大筋張力で跳躍に寄与できる配置にする必要がある。
また、実際の跳躍において重心軌道が正弦波である必要はなく、
小さな筋張力・仕事で跳躍が行える筋配置と制御を行うことが理想である。
そこで、動力学シミュレーションによる筋配置の検討を行う。

選択した筋配置で制御により跳躍が可能であるかどうかをシミュレーションにより確かめる。
物理エンジンにはMujocoを使用し、
胴体と脚からなる4リンクに加え、簡略化した尻尾を1リンクとして胴体につけた
2次元的なモデルにより跳躍を行わせた。
ホッピング動作の制御はRaibertらの制御\cite{Raibert}を基にする。
まず、立脚期においてはロボットの重心と足先に仮想的なばねの力となるような
足先力を定め、それを発揮できる関節トルクおよび筋張力を求める。
また、弾性のみではエネルギーの損失によりホッピングの振幅が減少するため、
RAMIELの制御器と同様の、
高さの維持のためのEnergy-Shaping Control\cite{RAMIEL}を行う。
また、滞空期においては、足先位置目標へと脚を動かす。
この位置目標は、立脚期に水平方向の速度を一定にするCG-print\cite{Raibert}と、
目標速度に対するフィードバックにより
決定する。

簡略化された筋配置をもとに、胴体へと延長し
多関節筋を利用した筋配置を\figref{figure:leg_use}に、
シミュレーションの様子を\figref{figure:mujoco_jump}に示す。
この筋配置は実験的に求めたものであり、数回の跳躍を可能としている。
筋張力を\figref{figure:mujoco_muscles}に示す。
この配置では、力が必要な部分を多関節筋により補うことで、
最大張力450Nで、4つの筋全てを活用して跳躍を可能としている。

以上で定まった筋配置を利用した脚の設計を\figref{figure:leg_design}に示す。
胴体部分に4つのワイヤモジュールを有し、
経由点によって脚へとワイヤを配線する。
}%

\section{Design of tail} \label{sec:tail}
\subsection{Musculoskeletal Structure of Living Kangaroo Tail}
\begin{figure}[t]
  \centering
  \includegraphics[width=1\columnwidth]{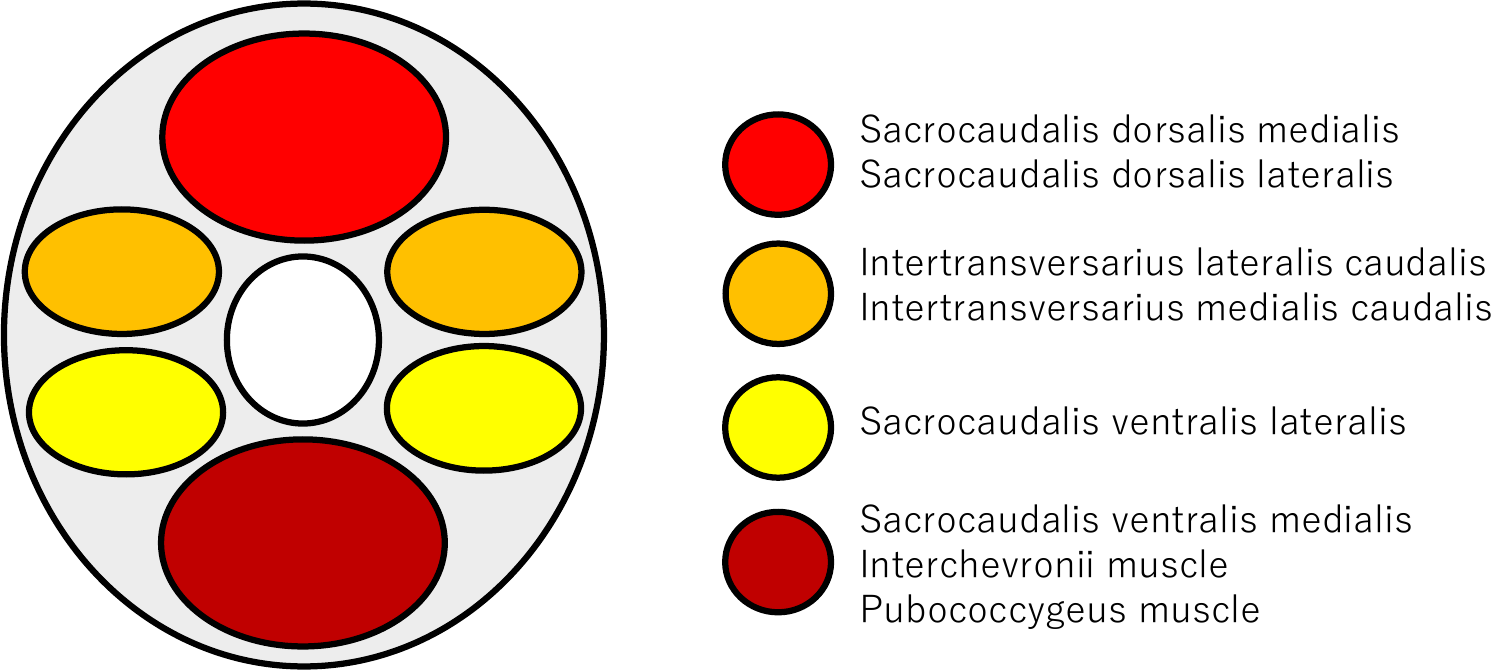}
  \vspace{-1.5ex}
  \caption{Muscle grouping in the cross-sectional view of tail.}
  \vspace{-1.5ex}
  \label{figure:tail_kangaroo}
\end{figure}

\begin{figure}[t]
  \centering
  \includegraphics[width=1\columnwidth]{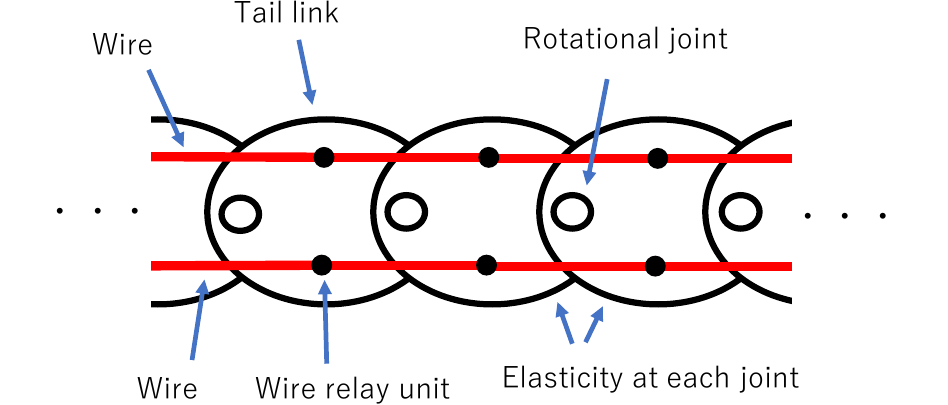}
  \vspace{-1.5ex}
  \caption{Simplified tail structure.}
  \label{figure:tail_structure}
  \vspace{-0.5ex}
\end{figure}

\switchlanguage%
{%
The musculoskeletal structure of the kangaroo's tail is shown in \figref{figure:tail_kangaroo} as a cross-sectional view of the tail, 
based on the anatomy\cite{Tailanatomy}. 
The caudal vertebrae, which form the skeleton of the tail, are smoothly connected from the spine, with approximately 20 links corresponding to the tail. 
Muscles extend from the base of the tail to the tip and are divided into four groups. 
The shape of the caudal vertebrae varies from the root to the tip of the tail. 
While the general arrangement of the muscle groups remains the same, the size of each muscle varies from the base to the tip. 
It enables a variety of movements.

The large number of joints in the tail makes it difficult to drive each joint independently. 
Therefore, we fabricate the tail by a simple, underactuated structure while incorporating the serial multi-jointed structure. 
The simplified tail structure is shown in \figref{figure:tail_structure}. 
Two wires are attached to the top and bottom of the tail, and the wires are routed in series through the relay points on each link from the root to the tip. 
The wires are pulled to swing up and down. 
Elastic elements and joint angle limits are provided at each joint. 
Even with a small number of wires, passivity of the springs can be used to perform the motion of an articulated structure.
}%
{%
カンガルーの尻尾の筋骨格構造について、解剖学\cite{Tailanatomy}に基づく尻尾の断面図を\figref{figure:tail_kangaroo}に示す。
尻尾の骨格を成す尾椎は背骨から滑らかに接続し、約20のリンクが尾に相当する。
筋肉は尾の根本から先端まで伸びており、4つのグループに分類される。
尾椎の形状は尾の根本から先端までで変化し、筋のグループの大まかな配置は変わらないものの
それぞれの筋の大きさが変化することで様々な運動を可能とする。

尻尾の関節数が多く、各関節を独立して
駆動することは困難である。
そこで、多関節の直列的な筋骨格構造を取り入れながら、簡略化した劣駆動構造によって尻尾を
製作する。
\figref{figure:tail_structure}に、簡略化した劣駆動構造を示す。
2本のワイヤを尻尾の上下に取り付けて、根本から先端までの各リンクに設けた経由点に直列にワイヤを通す。
ワイヤを引っ張って振り上げ・振り下げを行う。
また、各関節には弾性要素および関節角度限界を設ける。
少ないワイヤ数で受動性も利用して多関節構造の動作を行うことができる。
}%

\subsection{Verification of Tail Behavior of Articulated Elastic structures by Physics simulation}
\begin{figure}[t]
  \centering
  \includegraphics[width=1\columnwidth]{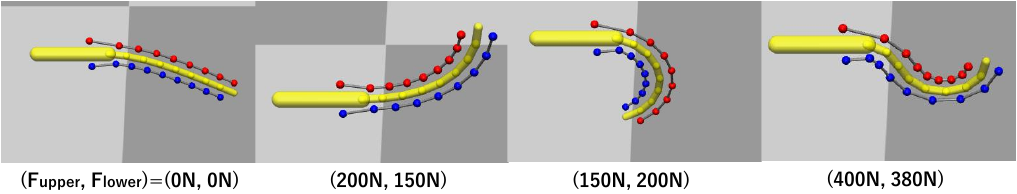}
  \vspace{-3ex}
  \caption{Tail simulation in mujoco.}
  \vspace{-3ex}
  \label{figure:tail_mujoco}
\end{figure}

\begin{figure}[t]
  \centering
  \includegraphics[width=1.0\columnwidth]{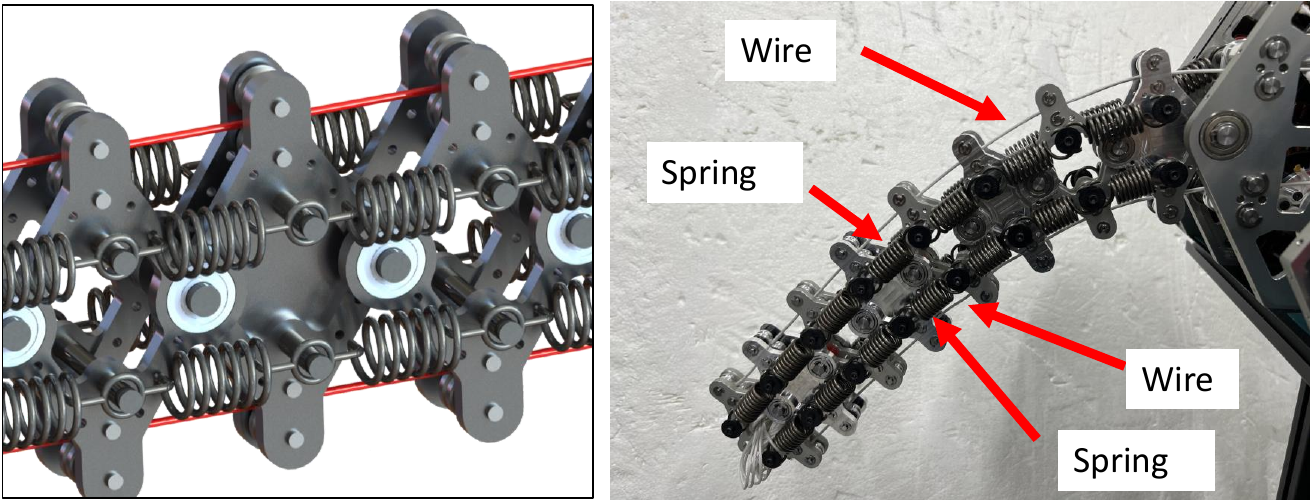}
  \vspace{-3ex}
  \caption{Mechanical design of the tail.}
  \vspace{-2ex}
  \label{figure:tail_hardware}
\end{figure}

\switchlanguage
{%
In realizing the tail with an articulated, elastic, underactuated structure, 
it is necessary to determine the indices of muscle performance, muscle arrangement, and elasticity strength that can be driven by wires. 
In this study, the same wire drive unit is used in the hind legs and tail to enable a large tension output in the tail. 
To confirm that the wire tension and elastic elements can be utilized in body support, 
the moment arm of wire and elasticity of joint will be determined by verifying the motion with dynamics simulation. 
The moment arm of the wire relative to the joint, the length of the link, and the strength of the elasticity are variables. 
Parameters were experimentally determined that allow the tail to remain stiffness at zero wire tension 
and to swing up and down with increasing stiffness by applying more tension to the wire.
The determined parameter is shown in \tabref{table:tail_table}.  
\figref{figure:tail_mujoco} shows the tail movement by this parameter. 
When the tension is set to 0, the tail forms a gentle curve due to the elastic elements. 
When the tension is increased, if either the upper or lower muscle tension is increased to a certain degree compared to the opposite side, 
the tail leans more toward the side where the muscle tension is stronger. 
On the other hand, if the upper and lower muscle tensions are close in value, 
the stiffness of the tail changes and the posture itself does not change significantly.

The mechanical design of the tail structure determined above is shown in \figref{figure:tail_hardware}. 
It utilizes tension springs as the elastic elements and has two wire modules in the torso. 
Each link of the tail is moved through the wires by means of upper and lower wire relay points.
}%
{%
前述した多関節で弾性を有する劣駆動構造の尾を実現するにあたり、
ワイヤによる駆動が可能な筋性能・筋配置・弾性の強さの指標を決定する必要がある。
今回、後脚と尻尾で同一のワイヤ駆動ユニットを使用して、
尻尾においても大きな張力を出力可能とする。
身体支持やバランスにおいてワイヤの張力と弾性要素を活用できることを確認するために、
動力学シミュレータにより動作を検証してワイヤのモーメントアームおよび弾性を決定する。
ワイヤの関節に対するモーメントアーム、リンクの長さ、弾性の強さを変数として、
ワイヤの張力が0の場合にも尻尾が剛性を保ちながら、
さらにワイヤに張力をかけることで剛性を大きくしたり上下に振ることが
可能であるパラメータを実験的に決定した。
\tabref{table:tail_table}に、決定したパラメータを示す。
また、\figref{figure:tail_mujoco}に、このパラメータによる尻尾動作を示す。
張力を0とした場合は弾性要素によって尻尾が緩やかなカーブを成す。
張力を大きくした場合、上下のいずれかの筋張力を逆側と比べてある程度大きくすると、
筋張力が強い側に尻尾が大きく寄る。
一方、上下の筋張力が近い値である場合、尻尾の剛性が変化して、姿勢そのものは大きく変化しない。

以上により決定した尻尾の構造を実装したものが\figref{figure:tail_hardware}である。
弾性要素として引張ばねを利用し、
胴体部分に2つのワイヤモジュールを有する。
尻尾の各リンクは上下の経由点によってワイヤを通して動かす。
}%

\begin{table}[tbhp]
 \begin{center}
  \caption{Physical parameters used in tail simulation}
  \begin{tabular}{cc} \toprule
  Parameter & Value \\ \midrule
  Tail total mass & 1.6 kg  \\
  Number of joints & 8 \\
  length of each link & 0.05 m \\
  wire moment arm & 0.035 m \\
  spring constant & 10 Nm/rad \\
  joint angle limit & $\pm{30}$ degree \\
  wire max tension & 450 N \\
  \bottomrule
  \end{tabular}
  \label{table:tail_table}
 \end{center}
\vspace{-2.5ex}
\end{table}

\section{Hardware of the kangaroo robot} \label{sec:hardware}
\subsection{High Power Wire Module Design}
\begin{figure}[t]
  \centering
  \includegraphics[width=1\columnwidth]{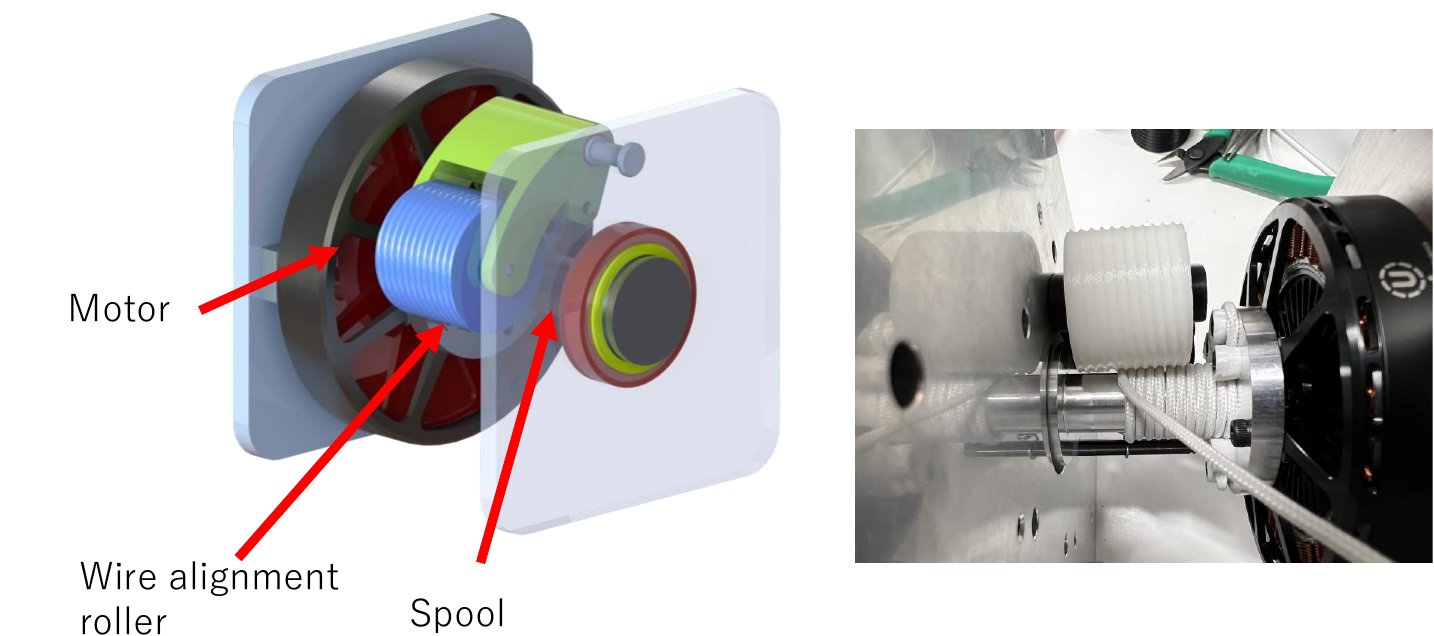}
  \vspace{-4ex}
  \caption{Design of the wire module.}
  \vspace{-3ex}
  \label{figure:wire_module}
\end{figure}

\begin{figure}[t]
  \centering
  \includegraphics[width=1\columnwidth]{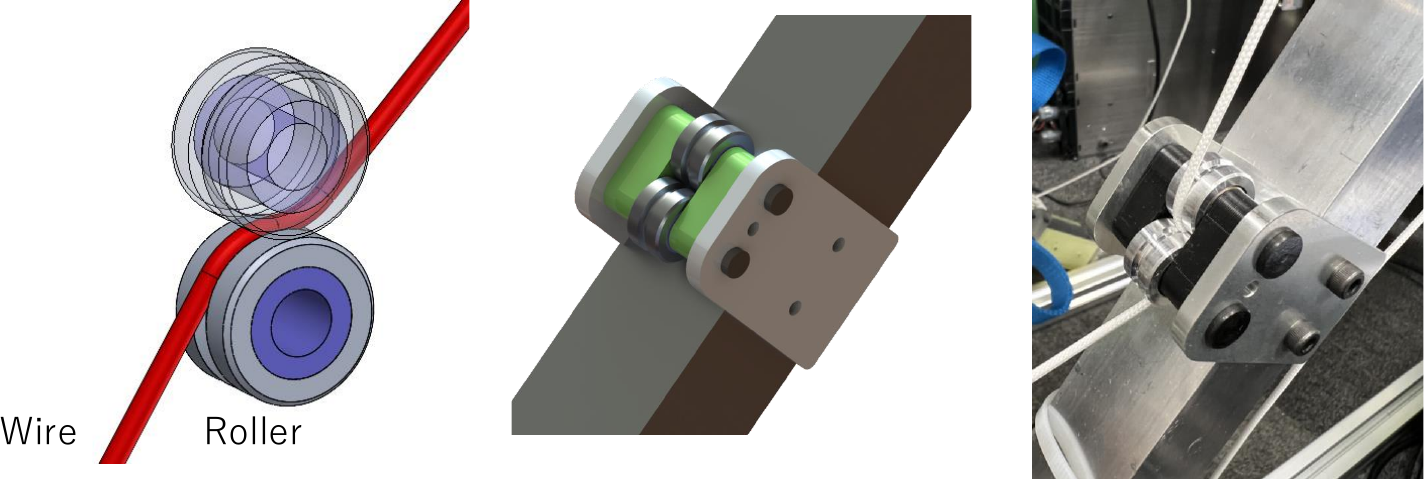}
  \vspace{-4ex}
  \caption{Design of the wire relay point.}
  \vspace{-1ex}
  \label{figure:relay_point}
\end{figure}

\switchlanguage%
{%
Wire-driven robots require a mechanism for winding and routing the wire. 
The winding pulley radius and routing structure can be freely determined and can be designed to exert the desired muscle tension and muscle speed.
In addition, backdrivability of the joints is high because actuators are not placed directly on the joints.

In order to further utilize the advantage of high backdrivability of the wire drive, 
we use a direct-drive system in which the reduction gear of the actuator is removed. 
This system is highly efficient and allows for flexible motion in response to external forces and the environment. 
On the other hand, to realize high output motion, the actuator must be able to generate large force without a reduction gear. 
To achieve both high backdrivability and high joint torque that enables jumping, 
we use a high-power motor driver that can apply 40A instantaneous current at 50V\cite{Ni} 
and a flat-type high-power brushless DC motor, T-Motor U8Lite L Kv 95 \cite{tmotor}.

Some wire-driven robots have load cells on the wire modules that have a large reduction ratio, to measure tension\cite{Musashi,MusashiW}.
On the other hand, with direct drive, it is possible to measure tension only by current feedback, 
but a mechanism is required to maintain a constant winding radius of the wire.
In addition, with wires that pass through many points, especially in multi-joint muscles, it is necessary to reduce the friction at the wire relay points.
Therefore, we developed a space-saving, lightweight, high-power wire module that can wind wires while aligning them. 
Also, we developed a high load-bearing and low-friction relay point using needle bearings. 
These are shown in \figref{figure:wire_module} and \figref{figure:relay_point}.

In the wire module, a cylindrical spool is attached to the rotating part of a flat type motor to wind the wire. 
Tension and speed can be adjusted by changing the spool radius. 
In the space around the spool, a roller is placed to align the wire and is pressed against the spool by a spring. 
The roller has a spiral groove dug into it. 
By making the radius of the roller twice the wire radius and the lead of the spiral twice the wire, 
the wire can be aligned and wound at a constant radius without tangling.
It can generate a maximum tension of 500N.

Next, at the relay point, a grooved pulley was press-fitted around the needle bearing and the wire was sandwiched between the two pulleys. 
This structure is compact and can withstand high tension, yet has low friction.
}%
{%
ワイヤ駆動ロボットは、ワイヤを巻き取る機構および配線する機構が必要となる。
巻き取りプーリー半径や配線構造を自由に定められ、
望みの筋張力や筋速度を発揮するように設計可能である。
また、関節に直接アクチュエータを配置しないため、
関節のバックドライバビリティが高い。

今回、ワイヤ駆動のバックドライバビリティの高さという長所をより活用するために、
アクチュエータの減速機を取り除いたダイレクトドライブ方式を利用する。
高効率で、外力や環境に対して柔軟に運動が可能である。
一方で、大出力の運動の実現のためには、アクチュエータが減速機無しでも大きな力を
発揮できる必要がある。
そこで、高いバックドライバビリティと、
跳躍を可能とする大きな関節トルクを両立して実現するために、
50Vで瞬間電流40Aを流すことが可能な大出力モータドライバ\cite{Ni}および、
フラット型の大出力ブラシレスDCモータであるT-Motor U8Lite L Kv95\cite{tmotor}を利用する。

ワイヤ駆動ロボットには、減速比の大きいワイヤモジュールで張力を測定するために、
モジュールにロードセルを搭載したロボットも存在する\cite{Musashi,MusashiW}。
一方、ダイレクトドライブにより、電流のフィードバックのみによって張力を計測できる。
但し、ワイヤを一定半径で巻き取る機構が必要である。
また、多関節筋をはじめとする、複数箇所に経由するワイヤに関しては、
経由点の摩擦を小さくする必要がある。
そこで、省スペース・軽量で、ワイヤを整列しながら巻き取ることが可能な
大出力ワイヤモジュールおよび、
ニードルベアリングによる高耐荷重・低摩擦な経由点をそれぞれ
\figref{figure:wire_module}、\figref{figure:relay_point}に示す。

ワイヤモジュールにおいて、フラット型モータの回転部に円柱のスプールを取り付けて
ワイヤを巻きつける。
スプール半径の変更により、張力と速度を調整できる。
また、スプールの周りの空間に、ワイヤを整列するローラーを設け、
ばねによってスプールへと押し付ける。
このローラーには螺旋状の溝が掘ってあり、ローラーの半径をワイヤ半径の2倍、
螺旋のリードをワイヤの2倍にすることで、
ワイヤを整列して、絡むことなく半径一定で巻き取ることができる。

次に、経由点では、小型で大きな張力に耐えられながら、摩擦の小さい構造として、
ニードルベアリングの周りに溝のついたプーリーを圧入し、2つのプーリーで
ワイヤを挟み込む方式をとった。
}%

\begin{figure}[t]
  \centering
  \includegraphics[width=1\columnwidth]{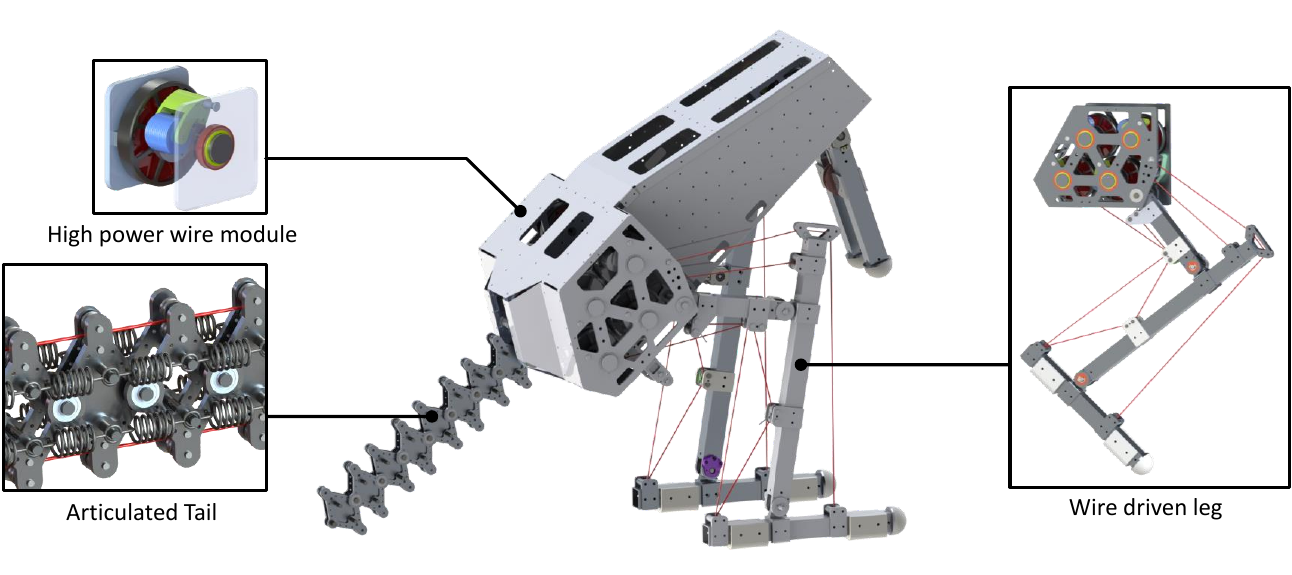}
  \vspace{-4ex}
  \caption{Fullbody design of the kangaroo robot.}
  \vspace{-4ex}
  \label{figure:kangaroo_cad}
\end{figure}

\subsection{Full Body Design of Kangaroo Robot}
\switchlanguage%
{%
The main physical parameters of the robot are shown in \tabref{table:kangaroo_table}. 
The overall design is shown in \figref{figure:kangaroo_cad}.
In addition to two hind legs and a tail, the robot has front legs for future use. 
11 motors are attached to the body in total. 
The torso is a monocoque structure made of sheet metal, with internal circuits such as batteries and motor drivers. 
Commands were sent to the robot by an external PC via a wire. 
The motor is powered by 50 V and each motor can carry a maximum current of 40 A. 
The maximum overall output is 20 kW.
}%
{%
製作したカンガルーロボットについて、主要なパラメータを\tabref{table:kangaroo_table}に示す。
また、写真は\figref{figure:kangaroo}に示した通りであり、
全体像を\figref{figure:kangaroo_cad}に示す。
2本の後脚、1本の尻尾に加えて、
将来的な活用を見据えて前脚を有し、
合計で11個のモータを搭載している。
胴体は板金によるモノコック構造とし、
内部にバッテリ、モータドライバ等の回路を有する。
モータの電源は50V、最大40Aまで電流を流すことができる。
また、最大全体出力20kWとした。
}%

\begin{table}[tbhp]
 \begin{center}
  \caption{Physical parameters of the kangaroo robot}
  \begin{tabular}{cc} \toprule
  Parameter & Value \\ \midrule
  Total mass & 18.5 kg  \\
  Torso length & 0.65 m \\
  Torso height & 0.19 m \\
  Torso width & 0.39 m \\
  Leg overall length & 0.68 m \\
  Tail length & 0.4 m \\
  Leg mass & 1.1 kg \\
  Tail mass & 1.6 kg \\
  \bottomrule
  \end{tabular}
  \label{table:kangaroo_table}
 \end{center}
\vspace{-1.5ex}
\end{table}

\section{Experiments} \label{sec:experiments}
\subsection{Jump Experiment} \label{subsec:jump}
\begin{figure}[t]
  \centering
  \includegraphics[width=1\columnwidth]{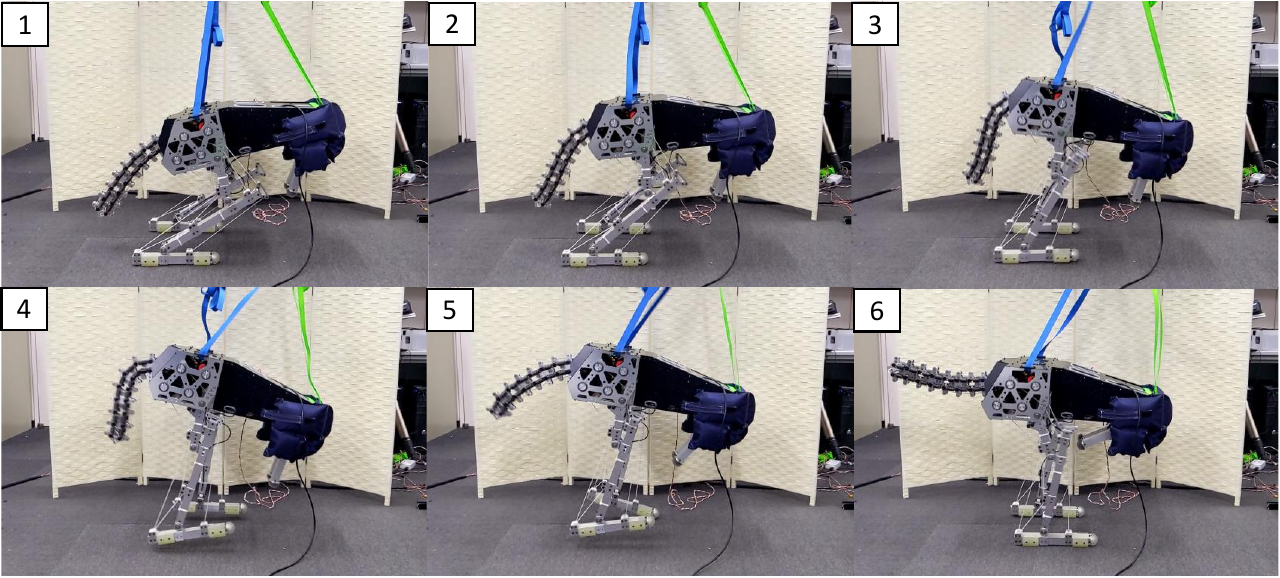}
  \vspace{-3ex}
  \caption{Jump experiment using legs.}
  \vspace{-1ex}
  \label{figure:jump}
\end{figure}

\begin{figure}[t]
  \centering
  \includegraphics[width=1\columnwidth]{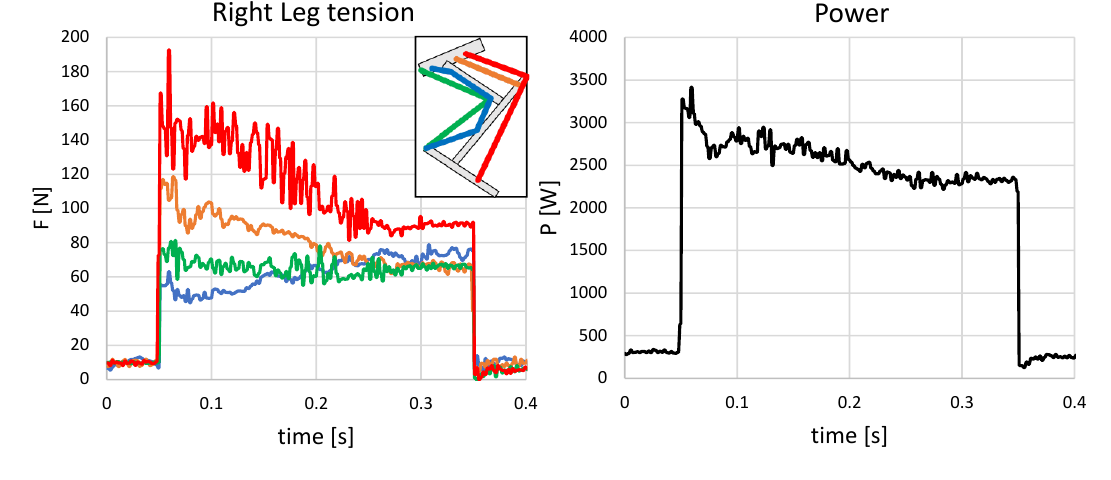}
  \vspace{-3ex}
  \caption{Leg muscle tension and Power in jump experiment.}
  \vspace{-4ex}
  \label{figure:jump_graph}
\end{figure}

\begin{figure}[t]
  \centering
  \includegraphics[width=1\columnwidth]{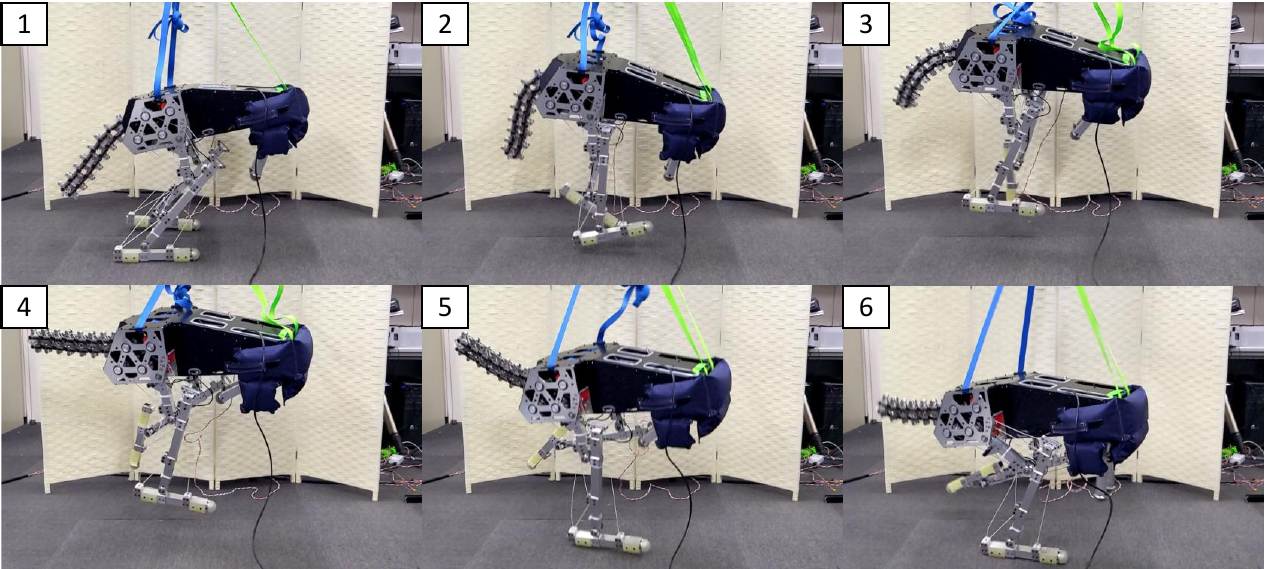}
  \vspace{-3ex}
  \caption{High jump experiment using legs. }
  \vspace{-1ex}
  \label{figure:jump_high}
\end{figure}

\begin{figure}[t]
  \centering
  \includegraphics[width=1\columnwidth]{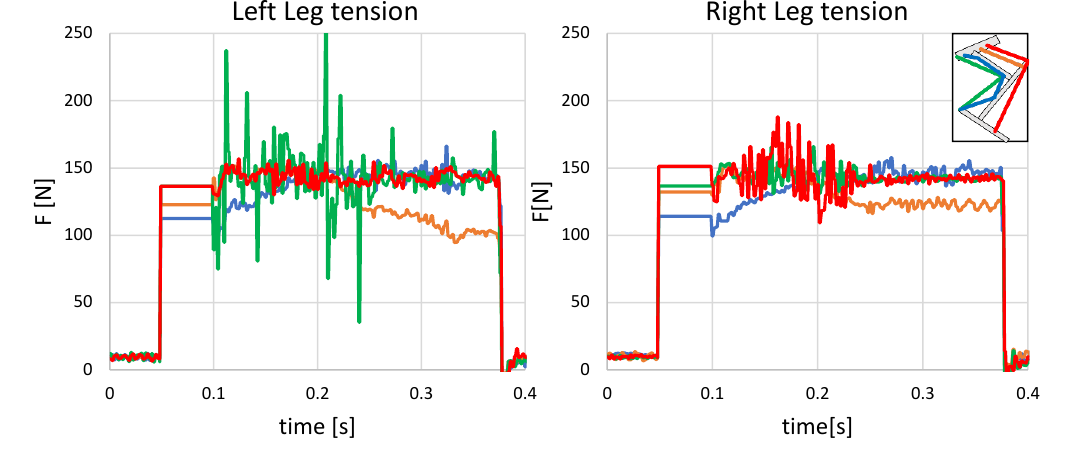}
  \vspace{-3ex}
  \caption{Leg muscle tension in high jump experiment.}
  \vspace{-4ex}
  \label{figure:jump_high_graph}
\end{figure}

\switchlanguage%
{%
First, we describe jumping experiments using the hind legs. 
The kangaroo robot is hung from above with belts attached to its waist and upper body. 
The torso is supported by the belts to keep it from falling. 
The torso is tilted at an angle of about 20 degrees from the horizontal. 
The robot is grounded only with its hind legs. 
The tail is not driven, and only the hind legs are used to make a single leap. 
Before the leap, all leg muscles are subject to a constant tension of 10N. 
In the leap, a constant torque is set at each joint. 
The muscle tension capable of exerting that torque is calculated and driven by the quadratic programming method according to the joint angles. 
After a certain time, at the end of the stance phase, the wire tension is returned to the initial constant tension.

\figref{figure:jump} shows the snapshots of the jump experiment.
The specified joint torques are 9.6 Nm, 19.2 Nm, and 1.4 Nm for the hip, knee, and ankle.
The maximum specified tension is 145 N.
The leg actuation duration was set to 0.3 seconds. 
The muscle tension calculated from the measured current values and the power of the entire two legs are shown in \figref{figure:jump_graph}. 
The hip part is pushed upward by 10 cm in a stable condition, and the sole of the foot is lifted off the ground by 3 cm.

In addition, the same joint torques are set and driven with the maximum specified tension of 190 N. 
The jumping motion is shown in \figref{figure:jump_high}, and the leg tension is shown in \figref{figure:jump_high_graph}. 
At the start of the jump, the muscle tension conversion is not performed correctly.
The impact during takeoff causes a delay in the circuit's signal, 
making it unable to apply the appropriate current to the motor based on the joint angle.
As a result, the muscle tensions in the left and right legs remain different for a long time, 
and the robot rotates to the right, but succeeds in jumping to lift the torso about 15 cm.

According to these results, we can see from the \figref{figure:jump_graph} that all four muscles of the leg contribute to the jump. 
In the design of the leg, to enable jumping at low maximum tension, 
the articulated muscles are used to increase the number of wires contributing to the joints that require large torques. 
As a result, even with wire drive that can exert force in only one direction, high output movement is possible by utilizing all muscles. 
Next, regarding the posture during jumping, the hip joints are not moving much as the leap begins with the hip joints almost fully extended.
If it jumps from a raised hip, it may be able to jump higher. 
Also, the specified torque of the ankle joint was smaller than that of the hip and knee. 
Because of the wire drive and the softness of the joint, the sole of the foot is in flexible contact with the ground. 
This causes the ankle joint to move in a direction where the lower leg and sole are vertical. 
On the other hand, with two legs, the robot is affected by the difference in the movement of the left and right leg. 
In particular, the relationship between muscle tension and joint torque changes with joint angle. 
Thus, when the angles of the two legs are different, the difference in the foot strength that can be exerted also becomes more pronounced.
}%
{%
まず、後脚を利用した跳躍実験について述べる。
カンガルーロボットの腰部および上半身にベルトをつけて上から吊るし、
ベルトで支持して倒れないようにした状態を保つ。
胴体は水平から20度程度の傾きとし、後脚のみで接地する。
尻尾は駆動せず、後脚のみで一回の跳躍を行う。
後脚の筋張力はすべて10N程度の一定張力の状態で上から吊るして跳躍を開始する。
跳躍においては、それぞれの関節に一定のトルクを設定し、
関節角度に応じて、そのトルクを発揮できる筋張力を二次計画法により計算して駆動する。
一定時間後、立脚期が終わり、ワイヤの張力を最初の一定張力へと戻す。

まず、\figref{figure:jump}に、最大指定張力を145Nとして、
関節トルクを根本から順に、脚が伸びる方向に9.6Nm, 19.2Nm, 1.4Nmとして、
筋張力に変換して駆動させた場合の跳躍の様子を示す。
脚の駆動時間は0.3秒とした。
また、計測した電流値から計算した筋張力、
2脚全体の電力を\figref{figure:jump_graph}に示す。
安定して腰部分を10cmほど上方向に押し上げ、
足裏は3cmほど地面から浮いている。

次に、最大指定張力を190Nとして、同様の関節トルクを設定して筋張力へと
変換して駆動した場合を\figref{figure:jump_high}、\figref{figure:jump_high_graph}に示す。
跳躍開始時に筋張力変換を正しく行えず、
左右の脚で筋張力が異なる状態が長く続いたために、ロボットが右方向へと回転してしまっているが、
胴体を15cmほど浮かせる跳躍に成功している。

跳躍の実験結果について考察する。
まず、最初の跳躍において、\figref{figure:jump_graph}より、脚の4つの全ての筋が跳躍に寄与していることがわかる。
脚の設計において、低い最大張力で跳躍を可能とするために、
多関節筋を利用して、大きなトルクが必要な関節に寄与するワイヤを増やしている。
結果として、一方向にしか力を発揮できないワイヤ駆動においても、
全ての筋肉を活用した大出力動作を可能にしている。
次に、跳躍時の姿勢について、股関節はほぼ伸ばした状態から跳躍を始めているため、あまり動いていない。
股関節を上げた状態から跳躍すれば、より高く跳躍できる可能性がある。
また、足首関節の指定トルクは股関節・膝と比べて小さくした。
ワイヤ拮抗駆動では関節が柔らかいため、
足裏と地面が柔軟に接して、下腿部と足裏が垂直になる方向へと足首関節が動いている。
一方、脚が2本あると、左右の動きの違いの影響を受ける。
特に、関節角度によって、筋張力と関節トルクの関係が変化するために、
両脚の角度が異なるときに、発揮できる足裏力の違いも顕著になる。
}%


\subsection{Tail Bending Motion Experiment}
\begin{figure}[t]
  \centering
  \includegraphics[width=1\columnwidth]{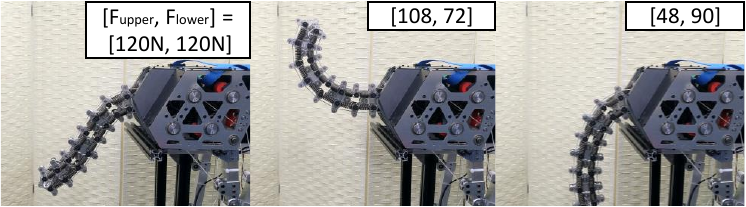}
  \vspace{-4ex}
  \caption{Tail bending experiment.}
  \vspace{-2ex}
  \label{figure:tail_bend}
\end{figure}

\switchlanguage%
{%
In the tail movement experiment, the body was fixed on a table and only the tail was moved to confirm the change in shape. 
The constant tension applied to the tail is shown in \figref{figure:tail_bend}. 
If either the tension of upper or lower muscle is larger than that of the opposite side, the tail bends in the direction of strong muscles. 
On the other hand, when the upper and lower muscle tensions are close, the shape of tail is similar to the case without muscle tension, but the stiffness of the entire tail changes. 
When the upper and lower muscle tensions are close and large, the shape of the tail becomes an S-shaped curve. 
Furthermore, when the muscle tension was set as a sinusoidal wave, the tail oscillated up and down, 
converging to a stable oscillation pattern when the upper and lower muscle tensions were set in opposite phases.

Based on the results of this experiment, we found that, firstly, the tail we fabricated is a little less elastic than in the simulation. 
This is because although elasticity is provided using a tensile spring, the moment arm changes depending on the angle, and the spring force weakens when the joint is bent greatly. Secondly, if the tension of the wire is set very high compared to the elasticity of the spring, the tail is fixed at the angle limitation of the joint.  
If the elasticity is made stronger, the wire and spring will provide higher stiffness, which may be advantageous in body support, but requires greater tension.
}%
{%
尻尾の動作実験においては、胴体を台の上に乗せて固定し、尻尾のみ動かして
形状の変化を確かめた。
\figref{figure:tail_bend}に、尻尾に一定張力をかけた様子を示す。
上下のどちらかの筋が、逆側と比べて大きな張力である場合、尻尾はその方向へと反る。
一方、上下の筋張力が近い値である場合、形状は筋張力なしの場合と近いが、
尻尾全体としての剛性が変化する。
また、上下の筋張力が近く、かつ大きな値である場合、S字カーブの形状となる。
さらに、筋張力として正弦波を出力した場合、
上下の筋張力を逆位相とすることで、
尻尾は上下に揺れ動き、安定した振動パターンに収束した。

尻尾の駆動について考察する。
まず、製作した尻尾はシミュレーションと比べて弾性が少し弱い。
引張ばねを利用して弾性を設けているが、角度によってモーメントアームが変化して、
関節を大きく折り曲げたときにばねの力が弱くなってしまうためである。
次に、ばねの弾性と比べてワイヤの張力をとても大きくすると、尻尾は関節角度限界で
固定されてしまう。
弾性をより強くすれば、ワイヤとばねにより高い剛性を発揮でき、
身体支持において有利になる可能性がある。
一方、より大きな張力が必要となる。
}%

\subsection{Motion Experiment Using Both Legs and Tail}
\begin{figure}[t]
  \centering
  \includegraphics[width=1\columnwidth]{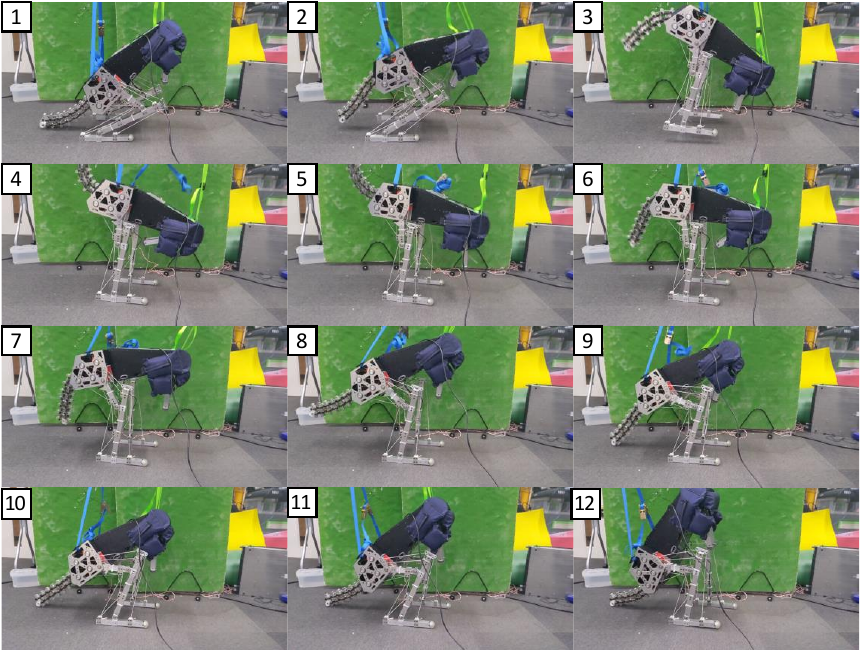}
  \vspace{-3ex}
  \caption{Motion experiment using both tail and legs.}
  \vspace{-3ex}
  \label{figure:tail_jump}
\end{figure}

\begin{figure}[tbh]
  \centering
  \vspace{-2ex}
  \includegraphics[width=1\columnwidth]{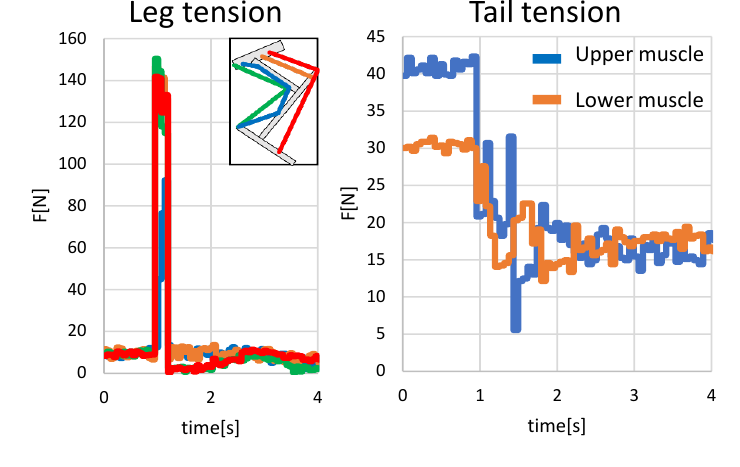}
  \vspace{-6ex}
  \caption{Muscle tension of legs and tail.}
  \vspace{-2.5ex}
  \label{figure:tail_jump_graph}
\end{figure}

\switchlanguage
{%
In the experiment for movements with the hind legs and tail, the torso is suspended by two belts and maintained in an inverted posture. 
The snapshots of jumping are shown in \figref{figure:tail_jump} and the corresponding muscle tension of hind legs and the tail are shown in \figref{figure:tail_jump_graph}. 
In the initial state, the tail is subjected to a tension of 40 N on the upper side and 30 N on the lower side to create a shape in which the tail is slightly bent upward. 
In this state, the whole weight of robot can be easily applied to the tail, leading the tail to be bent as shown in the first snapshot of \figref{figure:tail_jump}.  
Next, the robot leaps while swinging its tail up, with the muscle tension of hind legs increased and the stiffness of the tail reduced. 
In detail, hip joint torque of each hind leg is set as 9Nm, knee as 18Nm and ankle as 2Nm. 
These torques are converted to muscle tension, with the muscle tension limited to 145Nm. 
Meanwhile, the tail is returned to the extended posture with a target muscle tension of 25 N for both the upper and lower muscles of the tail. 
After 1 second, the designated tension of the tail is set as 15N, and the body weight is placed on the tail after landing. 
Due to the elasticity of the tail, the robot can sit down softly. 

Based on the result, 
we found that the legs and tail work well together in jump. 
The flexible tail and high power, backdrivable legs 
allow the robot to leap from a supported position by both tail and hind legs, using their flexibility while performing dynamic movements, 
and to soften the impact by their tails after landing.
}%
{%
後脚と尻尾を用いた動作の実験においては、胴体を2本のベルトで吊り下げ、倒立した姿勢を保つ。
最初に、尻尾の剛性を高くして、尻尾へともたれかかるような姿勢の状態にする。
次に、尻尾の張力を変化させるとともに、後脚に張力をかけることで尻尾を振り上げながら
跳躍する。
\figref{figure:tail_jump}に、尻尾にもたれかかった状態からの跳躍の様子を示す。
まず、初期状態で尻尾には上側40N、下側30Nの張力をかけて、尻尾を少し上へと反らせた形状にする。
この状態だと、尻尾は地面に対して丸い形状で接し、体重を尻尾へとかけやすい。
次に、後脚には最大指定張力を145Nとして、関節トルクを根本から順に、脚が伸びる方向に
9Nm, 18Nm, 2Nmとして、筋張力に変換して駆動する。
同時に、尻尾の筋の目標筋張力を上下とも25Nとして、尻尾を伸ばした状態へと戻す。
さらに1秒後に、尻尾の指定張力を15Nとして、着地後は尻尾へ体重をかける姿勢となる。
後脚および尻尾の筋張力を\figref{figure:tail_jump_graph}に示す。
尻尾の筋と弾性を利用した、尻尾と脚による3点接地状態から、
脚を利用して尻尾を上へと振り上げている。
また、着地後には尻尾の弾性により、やわらかく腰を下ろしている。

後脚と尻尾を利用した跳躍動作においては、これらの身体要素が相性良く作用することを確認できた。
多関節の尻尾は、弾性としての引張ばねと、2本のワイヤによって駆動し、柔軟な構造を有する。
脚は大出力でありながら、バックドライバビリティの高いワイヤモジュールおよび
拮抗ワイヤ駆動の活用により、柔軟な構造を実現している。
これにより、ダイナミックな動作を行いながら、柔軟さを活かして、
尻尾と後脚の両方による支持状態からの跳躍や、着地後に尻尾によって衝撃を
和らげることができている。
}%

\section{Conclusion} \label{sec:conclusion}
\switchlanguage%
{%
In this paper, we proposed a design method of a life-size wire-driven robot based on the kangaroo, 
which has unique musculoskeletal structures such as a powerful and flexible tail and legs.
Muscle arrangements of the legs were determined by analyzing muscle performance 
based on jumping trajectories and simulating jumping dynamics, using the musculoskeletal structure of the kangaroo as a reference. 
In addition, an articulated tail based on a biological structure, operated by wires and elasticity, 
was designed. 
We developed backdrivable high-power wire modules and relay points. 
Using them, we made the legs, tail, and whole body. 
Then, experiments were conducted on the legs and tail, respectively, and jumping and tail bending movements were performed.
Furthermore, a jumping motion from a three-point ground contact was realized using both the legs and the tail.
We confirmed that the large output and flexibility expected in the design were realized.

One future prospect is to use this kangaroo robot to perform complex movements seen in real kangaroos. 
For example, it is important to realize a five-legged walk using the tail and four legs, and a continuous leap using the tail for balance.
}%
{%
本研究では、力強く柔軟な尻尾や強力な脚といった固有の筋骨格構造を有するカンガルーを規範として、
等身大のワイヤ駆動ロボットを構成した。
生物の筋骨格構造を参考としながら、跳躍軌道に基づいた脚の筋性能分析、
ジャンプの動力学シミュレーションを行うことで筋配置を決定した。
これにより、生物の身体構造を規範としながらも等身大で大出力の脚を実現した。
さらに、生体構造を基にした、ワイヤと弾性により動作する多関節構造の尻尾について、
シミュレーションにより動作を確認しながら設計を行った。
ハードウェアの実装においては、ワイヤ駆動の良さを引き出すような、
バックドライバブルな大出力ワイヤモジュールおよび経由点を開発し、それらを利用した
脚、尻尾および全身を開発した。
そして、脚・尻尾それぞれについて、単体での実験を行い、跳躍および尻尾の曲げ動作を行った。
さらに、脚と尻尾の両方を利用した動作を実現した。
設計においては、生体構造規範と性能実現の双方を考慮していたが、
想定した大出力と柔軟さを実現できていることを実機により確認した。

将来の展望としては、このカンガルーロボットを利用して、生物のカンガルーにみられる
より複雑な動作を行わせることが挙げられる。
例えば、尻尾と4脚を利用した5足歩行や、
尻尾をバランスに利用した連続跳躍の実現が課題である。
}%

\vspace{-1ex}

{
  \bibliographystyle{IEEEtran}
  \bibliography{bib}
}

\end{document}